\documentclass[conference,table,xcdraw]{IEEEtran}

\usepackage[T1]{fontenc}
\usepackage{newtxtext}
\usepackage{newtxmath}

\usepackage{cite}

\usepackage{amsmath,amssymb,amsfonts}
\usepackage[ruled,vlined]{algorithm2e}
\usepackage{graphicx}
\usepackage{caption}
\usepackage{subcaption}
\usepackage{capt-of}
\usepackage{textcomp}
\usepackage{booktabs}
\usepackage{multirow}
\usepackage{colortbl}
\usepackage{xcolor}
\usepackage{xurl}
\usepackage{soul}
\usepackage[hidelinks]{hyperref}
\begin{document}

\title{DeepSeqCoco: A Robust Mobile Friendly Deep Learning Model for Detection of Diseases in Cocos nucifera }

\author{
    \IEEEauthorblockN{Miit Daga}
    \IEEEauthorblockA{\textit{Student, School of Computer Science} \\
    \textit{Engineering and Information Systems}\\
    VIT, Vellore, India \\
    miitdaga03@gmail.com\\
    ORCID: 0009-0005-4629-458X} 
\and
    \IEEEauthorblockN{Dhriti Parikh}
    \IEEEauthorblockA{\textit{Student, School of Computer Science} \\
    \textit{Engineering and Information Systems}\\
    VIT, Vellore, India \\
    dhriti.parikh2603@gmail.com\\
    ORCID: 0009-0008-5455-6676}
\and
    \IEEEauthorblockN{Swarna Priya Ramu}
    \IEEEauthorblockA{\textit{Professor, School of Computer Science} \\
    \textit{Engineering and Information Systems}\\
    VIT, Vellore, India \\
    swarnapriya.rm@vit.ac.in\\
    ORCID: 0000-0002-8287-9690} 
}
\maketitle

\begin{abstract}
Coconut tree diseases are a serious risk to agricultural yield, particularly in developing countries where conventional farming practices restrict early diagnosis and intervention. Current disease identification methods are manual, labor-intensive, and non-scalable. In response to these limitations, we come up with DeepSeqCoco, a deep learning based model for accurate and automatic disease identification from coconut tree images. The model was tested under various optimizer settings, such as SGD, Adam, and hybrid configurations, to identify the optimal balance between accuracy, minimization of loss, and computational cost. Results from experiments indicate that DeepSeqCoco can achieve as much as 99.5\% accuracy (achieving up to 5\% higher accuracy than existing models) with the hybrid SGD-Adam showing the lowest validation loss of 2.81\%. It also shows a drop of up to 18\% in training time and up to 85\% in prediction time for input images. The results point out the promise of the model to improve precision agriculture through an AI-based, scalable, and efficient disease monitoring system.
\end{abstract}

\begin{IEEEkeywords}
Precision Agriculture, Disease Detection, Deep Learning, Transfer Learning
\end{IEEEkeywords}

\section{Introduction}
Coconut tree is one such plant whose all parts, from the fruit to the trunk are useful in some or the other way. It is also known as the "tree of life" as it is used for food, medicine, fuel and so on.
However, coconut cultivation faces significant challenges in various countries around the world, most of which are either under developed or developing nations. In Côte d'Ivoire (a country in West Africa), an outbreak resulted in destruction of 350 hectares of coconut plantation and caused a loss of 12,000 tons of copra each year \cite{arocha2014identification}. In the South-east Asian country of Philippines, the second largest producer of coconut in the world, a 2012-2013 Philippine Coconut Authority (PCA) survey reported more than half a million Cadang-Cadang-infected palms in the country. According to their estimate, at a spread rate of 0.1-1\% annually, by 2032, around 3 million coconut trees would be at a risk of infection. If not intervened in timely, it could result in losses of approximately 300 million dollars by 2032 \cite{DOST2024}. India, the third largest coconut producer faces yield losses ranging from 10\% to 80\% due to diseases like Root Wilt Disease (RWD), Leaf Blight, Bud Rot and Basal Stem Rot \cite{esakkimuthu2020occurrence}.
In Sri Lanka, infestation due to white flies have led to a 40\% reduction in king coconut output with more than 4 million affected coconut palms \cite{FreshPlaza2023}.

To address these issues, the past decade has seen extensive advances in the field of smart agriculture, especially in precision agriculture \cite{balasundram2020precision}. The reason behind this is that even though there is a proportional decrease in the amount of agricultural land available, there is an increasing demand for food across the world due to an increased population. To handle this growing demand, the entire agricultural process should be mechanized. When using such a process mechanization as a remedy, the machines are made to perform most of the functionalities of the agricultural process there by increasing the effectiveness and yield \cite{sakthiprasad2019survey}. In the case of developing countries, the utilization of machines for agricultural processes is limited and inefficient, they instead focus on automating the entire functionalities of farming to improve the rural capabilities.

In order to address these challenges, modern day Artificial Intelligence (AI) powered solutions can be handy in providing real-time disease detection and help farmers especially in developing and under developed nations where age-old farming methods are still prevalent. This paper proposes a Deep Learning (DL) based mobile friendly framework \textbf{DeepSeqCoco} model, an intelligent solution to effectively identify coconut tree diseases through images. Utilizing powerful DL algorithms, it facilitates early and accurate diagnosis, enabling timely action and limiting crop loss, particularly in areas where conventional monitoring procedures are ineffective.
\section{Related Works}

Recent years have seen multiple researchers focus on societal problems and try to provide solutions, including agriculture as an important domain. The advancements in AI, Machine Learning (ML) and DL have paved the way for new methodologies in predicting the yield, pest detection, disease detection, and providing precautions to increase the production and hence the well-being of the farmers. This section discusses in detail the recent work carried out by various researchers around the world in the agriculture domain, especially, disease prediction in coconut trees. Anitha \textit{et al.} proposed a methodology for detecting the leaf disease in coconut trees using DenseNet 121 architecture. They performed pre-processing on the data set for handling the difference in color, orientation and shape of the leaves. ReLU activation function was utilized for enhancing the feature propagation. At the end of every epoch, a mechanism was used for automatically adjusting the learning rate. They validated their data set using the F1 score, accuracy, recall, and precision as evaluation parameters and achieved an accuracy of 99\% \cite{anitha2023disease}.

Bharathi \textit {et al.} came up with a methodology for early detection of diseases in coconut tree leaves using K-means clustering, statistical, Cubic Support Vector Machine (SVM) and  gray-level co-occurrence matrix (GLCM). While k-means clustering took care of segmentation, feature extraction was done through GLCM and Cubic SVM helped in classification. Data pre-processing was done using two main techniques: Contrast Enhancement and Histogram Equalization. While many features were extracted, such as mean, contrast, homogeneity, energy, entropy, variance, smoothness, standard deviation, and skewness; only mean, contrast, and energy were used with cubic SVM, achieving the highest accuracy of 97.3\% \cite{bharathi2020early}.

Nesarajan \textit{et al.} devised a model using image processing and DL techniques for prediction of nutrient deficiencies and pest diseases in coconut trees. They used SVM for classification of leaves and Convolutional Neural Networks (CNN) to predict the actual nutrient deficiency. While they tested multiple CNN architectures like EfficientNetB0, ResNet50 and VGG16, they finally decided upon EfficientNetB0 as it demonstrated the best accuracy of 93.72\% when compared to other architectures for prediction of nutrient deficiency. They also tested multiple algorithms including SVM, logistic regression and k-Nearest Neighbour (KNN) for pest detection but SVM outperformed all of them with an accuracy of 93.54\%. For this, they extracted features like color, shape and skewness from the images. K-means clustering technique was used for enhancing leaf image quality\cite{nesarajan2020coconut}.

Vidhanaarachchi \textit{et al.} developed a model for identifying coconut disease and pest infestations using DL architectures. They resized all the images in the dataset to a dimension of 300x300 and employed pixel normalization (dividing by 255) before using them with DL models like DenseNet-121, DenseNet-169, InceptionResNetV2, etc. They used Adam optimizer along with a learning rate ranging from 0.003 to 0.007 during the training phase. After testing multiple models, they finally settled with DenseNet-121 for Weligama Coconut Leaf Wilt Disease (WCWLD) detection as it gave 99\% training accuracy and 90\% testing accuracy. On the other hand, to determine the severity of symptoms associated with WCWLD, InceptionResNetV2 was used which achieved 99.69\% training accuracy and testing accuracy of 97\%. DenseNet-121 was also used for detecting Leaf Scorch Decline (LSD) which gave a training accuracy of 98.78\% and testing accuracy of 92\%. For detecting Magnesium (Mg) deficiency, they used the DenseNet-169 architecture which achieved 99.5\% and 88\%  training and testing accuracies respectively. Mask Region-based CNN (R-CNN) model with ResNet-101 was employed for Coconut Caterpillar Identification (CCI) along with Feature Pyramid Network (FPN) for feature extraction. This gave them a mean Average Precision (mAP) of 95.31\%. YOLOv5 object detection algorithm was used for counting caterpillars and gave a mAP of 96.87\%\cite{vidhanaarachchi2021deep}.

Tanwar \textit{et al.}, in their research, used a CNN-based model with a 7-layer architecture for the classification of coconut leaf diseases. They resized the images to a standard size of 224x224 and standardized brightness and density as part of image pre-processing. The convolution layer parameters included a kernel size of 4x4 along with 124 filters, as well as padding and a stride of 2 each. The pooling layers used max or average pooling to reduce spatial dimensions while also extracting key features. The final dense layer used SoftMax activation function for classification. For optimization purposes, Back-propagation was used for error minimization during training. Forward and backward passes were used to adjust weights with every iteration. They validated their data set on F1 score, accuracy, recall, and precision as evaluation parameters and achieved 99\% overall training accuracy and a validation accuracy of 98\% \cite{tanwar2023intelligent}.

Singh \textit{et al.} made use of Thresholding, Watershed, and K-means Clustering Segmentation (KMC) to pre-process images. The best validation accuracy of 85.23\% was achieved on making use of KMC. Then to enhance the dataset, they augmented the images using various rescaling methods, flips, shears, as well as zooms and expanded  it to 39,000 samples. It was then tested with a custom made CNN model with varying layers. The highest validation accuracy of 96.94\% and a Cohen’s Kappa of 0.91 was achieved when a 4-layer architecture was used, demonstrating the effectiveness of combining KMC and an optimized CNN for a high classification performance\cite{singh2021disease}.

Thite \textit{et al.} used IrfanView software to resize and enhance the photos that were taken with the rear camera of the Samsung F23 5G Mobile. Three DL architectures were then used to analyze the dataset: VGG16, ResNet50, and MobileNetV2. After training, ResNet50 achieved the highest accuracy of 94\%, compared to its initial accuracy of 0.4\%. In order to validate the models, five-fold cross validation was used and ResNet50 performed better than the others with 92.25\% precision, 93.42\% recall, and 93.88\% F1-score.\cite{thite2023coconut}.

Brar \textit{et al.} relied on the ResNet50 DL architecture. They improved upon the model images with augmentation techniques, including random rotation and flipping, before training. The images were organized in a 70 and 30 percent split for training and testing. The team used Stochastic Gradient Descent (SGD) optimization to train the model for 50 epochs, attaining a striking 90.77\% accuracy. Their method was especially accurate in discerning important cases at 97.76\% accuracy.\cite{brar2023smart}.

Kaur \textit{et al.} approach consisted of using a fine-tuned ResNet50 model for multi-class classification of coconut leaf diseases using both Adam and SGD optimizers. They expanded their initial dataset of 1000 images by using contrastive and zoom pre-processing techniques, enabling them to enhance it to 10,222 images. Multiple epochs (10-50) were run over the model with 40 epochs producing the best results. The highest accuracy of 98.70\% was achieved with the ResNet50 architecture using Adan optimizer, followed by 97.90\% with SGD. The model also performed strongly, achieving F1-scores of 0.9562 and 0.942 for Adam and SGD respectively at epoch 40, demonstrating the power of DL techniques for coconut leaf disease classification.\cite{kaur2024multi}.

Kavitha \textit{et al.} employed faster R-CNN with backbone ResNet-50 architecture to reduce over-fitting while detecting and classifying diseases in coconuts. The authors filtered the data and applied normalization with feature extraction using VGG-16 with max pooling layers. For classification, they used Faster R-CNN with a backbone of ResNet-50, applying the ResNet-50 model that was pre-trained on ImageNet to capture high-level features and counter the vanishing gradient effect. This method outperformed previous approaches that used CNN (97.45\%), DCNN (98.1\%) and LSTM (97.2\%) by receiving an accuracy of 98.4\%, precision of 97.1\%, and recall of 96.8\%. \cite{kavitha2024cocos}.
The summary and further details of the related works are given in Table \ref{tab:summary}.

In spite of the several advances in coconut tree disease prediction (or detection) using techniques like DL, ML and image analysis, there are a few limitations. The existing methods rely on complex frameworks that require ample computational power. This makes them unsuitable for real time deployment on devices like mobile phones. Additionally, a few methods like the ones depending on conventional ML (e.g., SVMs, KMC, etc) require extensive manual work to create features which might end up not even working well with different datasets. Moreover, although CNN-based architectures have shown high accuracy in controlled environments, their reliability might not be up to the mark when tested in real world agricultural conditions where different factors such as lighting conditions and environmental noise pose a significant challenge.
To address these challenges, we propose \textbf{DeepSeqCoco}, a robust and mobile-friendly DL model designed for high accuracy and efficient computation. The model uses sequential DL techniques to better the feature extraction and improve classification tasks. By using a carefully optimized architecture, an augmented dataset and flexible learning rate strategies, our model makes sure to provide a reliable output for disease detection with the use of minimal resources. This makes it a scalable and practical solution for farmers all around the world.

\begin{table*}[!h]
\caption{Summary of Related Works}
\label{tab:summary}
\begin{tabular}{|c|l|l|l|c|l|}
\hline
\rowcolor[HTML]{EFEFEF} 
\textbf{References}            & \multicolumn{1}{c|}{\cellcolor[HTML]{EFEFEF}\textbf{\begin{tabular}[c]{@{}c@{}}Methodology\\ Used\end{tabular}}}         & \multicolumn{1}{c|}{\cellcolor[HTML]{EFEFEF}\textbf{\begin{tabular}[c]{@{}c@{}}Data Set \\ Used\end{tabular}}}                                                                 & \multicolumn{1}{c|}{\cellcolor[HTML]{EFEFEF}\textbf{Classified Classes}}                                                                                                                                                                                                                                                                                                                 & \textbf{\begin{tabular}[c]{@{}c@{}}No.of\\ Epochs\end{tabular}}               & \multicolumn{1}{c|}{\cellcolor[HTML]{EFEFEF}\textbf{Limitations}}                                                                                                                                                                    \\ \hline
\cite{anitha2023disease}                        & DenseNet 121                                                                                                             & \begin{tabular}[c]{@{}l@{}}Hand collected 529 \\ images using smart \\ phone\end{tabular}                                                                                      & \begin{tabular}[c]{@{}l@{}}* Yellowing\\ * Caterpillars\\ * Flaccidity\\ * Drying of leaflets\\ * Healthy\end{tabular}                                                                                                                                                                                                                                                                   & 20                                                                            & \begin{tabular}[c]{@{}l@{}}* Environmental variations like\\    leaf textures not validated\\ * Scalability issues\\ * Very small data set\\ * Manual cropping used\end{tabular}                                                     \\ \hline
\cite{bharathi2020early}                   & \begin{tabular}[c]{@{}l@{}}SVM + GLCM \\ + k-means \\ Clustering\end{tabular}                                            & Personal data set                                                                                                                                                              & \begin{tabular}[c]{@{}l@{}}* Leaf rot\\ * Leaf spot\\ * Leaf blight\\ * Healthy\end{tabular}                                                                                                                                                                                                                                                                                             & -                                                                             & \begin{tabular}[c]{@{}l@{}}* No meta data about the data set\\ * No clear picture about how the \\    model would perform on larger \\    data sets.\end{tabular}                                                                    \\ \hline
\cite{nesarajan2020coconut}                       & \begin{tabular}[c]{@{}l@{}}SVM + \\ EfficientNet B0 \\ + k-means \\ Clustering\end{tabular}                              & \begin{tabular}[c]{@{}l@{}}Digital camera \\ captured personal \\ data set\end{tabular}                                                                                        & \begin{tabular}[c]{@{}l@{}}* Healthy\\ * Nutrient deficient \\   (Potassium, Nitrogen,\\    Magnesium)\\ * Pest affected\end{tabular}                                                                                                                                                                                                                                                    & -                                                                             & \begin{tabular}[c]{@{}l@{}}* No meta data about the data set\\ * Performance of EfficientNet B0\\    depends on computational\\    resources\\ * Lack of soil and environmental\\    considerations\end{tabular}                     \\ \hline
\cite{vidhanaarachchi2021deep}                        & \begin{tabular}[c]{@{}l@{}}DenseNet 121\\ DenseNet 169\\ InceptionResNet V2\\ ResNet 101 \\ YOLOv5\end{tabular}          & \begin{tabular}[c]{@{}l@{}}Pictures collected \\ by skilled employees \\ from the Coconut\\ Research Institute of\\ Sri Lanka (CRISL) \\ and crowdsourced \\ photographs\end{tabular} & \begin{tabular}[c]{@{}l@{}}Weligama Coconut Leaf \\ Wilt Disease\\ * 4 classes (Healthy, Flaccidity,\\    Uneven Yellowing, Drying of \\    Leaflets\\ CCI (Coconut Caterpillar Infestation)\\ * 3 classes (With CCI, Without CCI,\\    Background)\\ LSD (Leaf Scorch Decline)\\ * binary class (Healthy, Infected)\\ Mg Deficiency\\ * binary class ( Healthy, Deficient)\end{tabular} & \begin{tabular}[c]{@{}c@{}}100 \\ for \\ YOLOv5\end{tabular}                  & \begin{tabular}[c]{@{}l@{}}* Environmental variations like\\    lighting and leaf appearance not\\    validated\\ * Scalability issues\\ * Crowd sourced data set can\\    introduce inaccuracies\end{tabular}                       \\ \hline
\cite{tanwar2023intelligent}                       & \begin{tabular}[c]{@{}l@{}}CNN based model \\ with 7 layer \\ architecture\end{tabular}                                  & \begin{tabular}[c]{@{}l@{}}550 images from \\ Kaggle\end{tabular}                                                                                                              & \begin{tabular}[c]{@{}l@{}}* Yellowing\\ * Caterpillars\\ * Flaccidity\\ * Drying of leaflets\\ * Healthy\end{tabular}                                                                                                                                                                                                                                                                   & 20                                                                            & \begin{tabular}[c]{@{}l@{}}* Very small data set\\ * Scalability issues\\ * Limited diversity in disease\\    stages\\ * Need for continuous parameter\\    tuning\end{tabular}                                                      \\ \hline
\cite{singh2021disease}                       & \begin{tabular}[c]{@{}l@{}}2D CNN model, \\ VGG16, VGG19, \\ MobileNet,\\ DenseNet201, \\ InceptionResNetV2\end{tabular} & \begin{tabular}[c]{@{}l@{}}Self collected \\ coconut leaf data \\ set\end{tabular}                                                                                             & \begin{tabular}[c]{@{}l@{}}* Healthy\\ * Stem Bleeding\\ * Pest infection\\ * Leaf Blight\end{tabular}                                                                                                                                                                                                                                                                                   & \begin{tabular}[c]{@{}c@{}}150 initially\\ 50 for fine \\ tuning\end{tabular} & \begin{tabular}[c]{@{}l@{}}* Limited data set\\ * Overfitting due to pre-trained\\    models\\ * Computational resource \\    constraints\end{tabular}                                                                               \\ \hline
\cite{thite2023coconut}                       & \begin{tabular}[c]{@{}l@{}}VGG16, ResNet50, \\ MobileNetV2 + \\ Transfer Learning\end{tabular}                           & \begin{tabular}[c]{@{}l@{}}Mendeley Data : \\ Coconut Tree \\ Disease Data set\\ (5,798 images)\end{tabular}                                                                   & \begin{tabular}[c]{@{}l@{}}* Bud rot\\ * Gray leaf spot\\ * Leaf rot\\ * Stem bleeding\end{tabular}                                                                                                                                                                                                                                                                                      & -                                                                             & \begin{tabular}[c]{@{}l@{}}* Data set collected from specific \\    region (Kendur, Maharashtra)\\ * signs of overfitting shown during \\    testing\\ * Performance gap between \\    training and test sets\end{tabular}           \\ \hline
\cite{brar2023smart}                        & \begin{tabular}[c]{@{}l@{}}ResNet50 DL model +\\ SDG optimizer\end{tabular}                                              & \begin{tabular}[c]{@{}l@{}}10,000 photographs \\ of coconut trees'\\ leaves collected \\ from plantations,\\ research institutes\\ and public data \\ sets\end{tabular}                    & \begin{tabular}[c]{@{}l@{}}* Healthy leaves\\ * Mild level\\ * Moderate level\\ * Severe level\\ * Very severe level\\ * Critical level\end{tabular}                                                                                                                                                                                                                                     & 50                                                                            & \begin{tabular}[c]{@{}l@{}}* False positive rates need to be \\    addressed\\ * Data set diversity needs to be\\    improved\\ * Performance varies between \\    severity levels\end{tabular}                                      \\ \hline
\cite{kaur2024multi}                      & \begin{tabular}[c]{@{}l@{}}ResNet50 with \\ Adam and\\ SGD optimizer\end{tabular}                                        & \begin{tabular}[c]{@{}l@{}}Kaggle data set \\ of coconut\\ leaf images \\ (10,222 images\\ after augmentation)\end{tabular}                                                    & \begin{tabular}[c]{@{}l@{}}* Drying of leaves\\ * Caterpillars\\ * Leaflets\\ * Yellowing\\ * Flaccidity\end{tabular}                                                                                                                                                                                                                                                                    & 40                                                                            & \begin{tabular}[c]{@{}l@{}}* Model performance is highly\\    dependent on epoch selection\\ * Requires high quality clear \\    images\\ * Limited to visual disease\\    identification\\ * 
   Computationally intensive\end{tabular} \\ \hline

\cite{kavitha2024cocos} & 
\begin{tabular}[c]{@{}l@{}}Faster RCNN \\ with ResNet50\\ \& VGG16\end{tabular} & 
\begin{tabular}[c]{@{}l@{}}Coconut disease \\ dataset with 5,015 \\ images\end{tabular} & 
\begin{tabular}[c]{@{}l@{}}* Bud rot\\ * Gray leaf spot\\ * Stem bleeding\\ * Healthy\end{tabular} & 
- & 
\begin{tabular}[c]{@{}l@{}}* Complex\\ * Time intensive processing\\ * Potential overfitting\end{tabular} 
\\ \hline 

\end{tabular}
\end{table*}

\section{Background and Fundamentals}
This section discusses in detail the various fundamentals needed for better understanding of the proposed framework. The traditional DL, TL models and optimizers currently available in literature, their categories, advantages, disadvantages and applications are discussed in subsections. 
\subsection{Deep Learning}
Traditional ML systems faced significant challenges when processing raw data like speech, text, images, etc. directly. Developers had to create special tools to convert unprocessed data into a form that computers could understand and decode. For example, in the case of images, developers had to write programs that would first extract important details like edges, shapes and patterns and then explain it to the computer for it to learn. The computer would then use these features to make predictions about new images. This used to increase the work load of the people while also increasing the time required for development. It required extensive knowledge of the domain which was often difficult to obtain.To solve this problem, DL emerged as an excellent solution.
DL is a specialised subfield of ML which is inspired by how the human brain works. It leverages aritificial neural networks for learning patterns and relationships between different sections of the input \cite{sarker2021deep}. Neural networks are a type of computer models used to recognize patterns in data. It is made by connecting nodes, which work like the human brain. Every node takes in information, tries to learn from it and passes it to the next node or layer. It uses a technique called Backpropagation which is like a feedback system. In this, the network makes a prediction and then compares it to the correct answer to check for error. If an error is detected the network works in the reverse direction and try to adjust weights between the nodes or neurons so that the next time when a similar information has to be processed, it is able to make a better prediction. The adjustments are made through tiny tweaks to the connection weights between the neurons.

There are different types of DL models namely: Deep belief networks (DBNs), Recurrent neural networks (RNNs), CNNs, Deep reinforcement learning (DRL) and Autoencoders which are described below.
\begin{enumerate}
    \item CNNs: These are particularly designed for image processing and recognition tasks as they excel in detecting patterns in images such as shapes and textures. This helps it in identifying objects. CNNs consist of multiple layers:
    \begin{enumerate}
        \item Convolutional layers: These layers help the network by applying filters i.e small matrices to scan the image in order to find patterns. Every matrix is used to detect different features like edges, corners, etc.
        \item Pooling layers: These layers are used to optimize the size of data (called downsampling). It makes sure that important information is not lost in the process. This makes the model more efficient.
        \item Fully connected layers (FCL): The extracted features, after passing through several convolutional and pooling layers are flattened. FCL then makes a classification decision on the basis of the flattened features.
    \end{enumerate}
    \item RNNs: These are normally used for processing sequential data which makes it suitable for workloads that include understanding patterns as time passes. It is ideal for tasks like Natural Language Processing (NLP) and Speech recognition. Compared to CNNs, which analyzes data all at once, RNNs store information from older inputs using hidden underlying states.
    \item DRL: It is a combination of DL and reinforcement learning (RL). In RL, a system learns by interacting with its surrounding. It uses a feedback system for learning i.e, it gets rewards for performing desired action and penalties for the undesired ones. It does not use labeled data for this purpose and works on the principle of maximizing the long term reward. Use cases of DRL include game playing and autonomous vehicles.
    \item DBNs: These are specifically useful for unsupervised learning tasks. It learns useful features from data without the presence of labeled data. Its working is based on arranging multiple layers of random and hidden factors (called latent variables) that help the model to understand intricate information. Through the usage of this, the DBN can find useful representation of data which is otherwise hidden. The "belief" part of the name comes from the fact that they use a probababilistic approach to form a "belief" about the data after passing through multiple layers.
    \item Autoencoders: These are unsupervised learning algorithms that are employed for feature learning, denoising, and data compression. Autoencoders are composed of an encoder, which is used to compress the input to a lower dimensional vector, and a decoder, which produces the original input from the lower dimensional representation. The goal of its training is to learn to compress data with a minimum reconstruction error. There are 3 types of Autoencoders : Denoising Autoencoder which, as the name suggests, is used to remove noise from data by learning clear representations,VAE (Variational Autoencoder) which generates new data using a learned distribution and Sparse Autoencoder which is used to force most neurons to become inactive in order to improve feature extraction in an efficient manner.
\end{enumerate}

Even though there are many advantages of DL, there are a few challenges to it too. First is that its data requirement is very high i.e, it requires vast datasets which can be difficult and costly to obtain. Another challenge is that DL requires training on high performance Graphics Processing Units (GPUs) or Tensor Processing Units (TPUs), depending on the application and dataset size, which are often priced very high and are thus out of reach for individuals.
\subsection{Transfer Learning}
In Conventional ML methods it is necessary for the models to be trained from the ground up using labeled data that not only has the same feature space but also shares the same distribution as the task the model that it is ultimately intended to solve. Yet, in numerous real-world applications, it is not possible to gather together the large number of high-quality labeled instances that are required in order to train a good model. This is where Transfer learning comes into the picture. It enables us to use the knowledge that we gain from a related source domain and apply it to another target domain even when the two of them have completely different feature spaces \& distributions.

This method has applications in numerous areas. When it comes to computer vision, the DL models that serve as the foundation for ResNet and VGG are pretrained on huge datasets. These are subsequently fine-tuned for the particular images and tasks that one is likely to encounter in the domain of medical image analysis.NLP benefits from the same kind of modeling: BERT and GPT serve as platforms for information retrieval and question-answering systems. These models also perform sentiment analysis pretty well. So we can say that, indeed, model adaptation to the extremely variable circumstances one can encounter in the real world is core to many efforts in NLP. Robust autonomous systems, including intelligent robots and self-driving cars, are built using similar principles. Furthermore it can be utilized in the financial sector for creating fraud detection models by using  archival transaction data to detect new fraudulent activites.

Various methods can be used to carry out transfer learning. Some of the more commonly used methods include the following:

\begin{itemize}
    \item Feature Extraction: Transfer learning is carried out using a pre-trained model to derive features from the source and target domains.
    \item Fine-tuning: The process of retraining the last few layers of a deep network that have been pre-trained on a large dataset (the source domain) so that the model can better handle a new but related task (the target domain).
    \item Instance-based Transfer: Transfer learning using re-sampling of the target domain and re-weighting of instances taken from the source domain.
\end{itemize}

Broadly, transfer learning can be classified into three main categories.\cite{pan2009survey} 

\begin{enumerate}
\item Inductive transfer learning helps us typically in dealing with scenarios where the source and target tasks differ but the underlying knowledge is shared between them. This is very commonly seen in multi-task learning, where the labelled data is available for the target domain. For example, when we perform sentiment classification for different products, labelled reviews from one product category is utilized to help us train the model for another category.

\item Transductive transfer learning, is generally applied when the source and target tasks are the same, but their feature distributions are varied. For example, adapting a spam detection model that has been trained on English emails to classify spam in French emails.

\item Unsupervised transfer learning is applied when both the source and target tasks do not have any labeled data, which makes it particularly useful for applications like clustering and dimensionality reduction. It leverages patterns and structures that it learns from a large and diverse dataset to enhance its performance on a smaller and related dataset. For example, a model desisgned and trained to identify the underlying patterns in a large collection of images can be adapted to improve clustering in a smaller and a more domain-specific dataset without requiring explicit labels. 
\end{enumerate}

The major advantage of using  transfer learning is that it significantly reduces the dependency that we have on collecting large labeled datasets. Using it we can leverage the knowledge gained from related domains on the target domain greatly reducing the computational costs which would have otherwise been required to train a model from scratch. Another benefit is that, pretrained models converge faster making the adaption to newer tasks quicker.

However transfer learning is not without its own challenges. One of the biggest issues that we could face includes, negative learning- which occurs when knowledge gained from an unrelated source domain degrades the performance of the model on the target domain rather than improving it and domain shift- where the differences in the feature distributions between the source and target data make adaptation harder.

A key point to keep in mind when using transfer learning is Model selection, as choosing the appropriate pretrained model as well as the transfer strategy is very crucial for achieving optimal results.

\section{Materials and Methods}
This section introduces the transfer learning models like EfficientNet, VGGNet, MobileNet and ResNet. Each subsection discusses the structure of the TL model, the advantage as well as challenges in their usage. 
\subsection{EfficientNet}
EfficientNet is a family of CNNs launched by Google Brain in 2019 and is specially designed for image classification tasks. It is devised in such a way that it uses lower number of parameters and lesser computational resources compared to VGG and ResNet to give a high degree of accuracy \cite{tan2019efficientnet}.     There are 8 versions of EfficientNet which are measured in terms of Floating Point Operations Per Second (FLOPS) (which measures the workload required for the model to process an input). Their classification accuracy is measured using two main metrics: Top-1 accuracy percent and Top-5 accuracy percent.
\begin{enumerate}
    \item Top-1 accuracy: This shows the percentage of times the network's highest confidence prediction is correct. For example, if an image of a cat is given to the model and if the model classifies it as cat, then it counts as a correct prediction.
    \item Top-5 accuracy: This shows the percentage of times the network's top 5 predictions contains the correct class for the input image. It is useful in cases where multiple class images may look similar to each other. For example, if images of different breeds of dog are given to the model, and if the model's top 5 prediction contains the correct answer for the image, then it is counted for calculating the top 5 accuracy. Further details about the different versions is given in Table \ref{tab:efficientnet}.
\end{enumerate}
\begin{table}[htbp] 
\centering
\caption{EfficientNet : Versions, Parameters, Size and Accuracy}
\label{tab:efficientnet}
\footnotesize 
\setlength{\tabcolsep}{3pt} 
\begin{tabular}{|c|c|c|c|c|c|}
\hline
\rowcolor[HTML]{EFEFEF}
\textbf{Model} & \textbf{\begin{tabular}[c]{@{}c@{}}Parameters\\  in \\millions \\ (M)\end{tabular}} & \textbf{\begin{tabular}[c]{@{}c@{}}FLOPs \\ in \\ Billions\\ (B)\end{tabular}} & \textbf{Input Size} & \textbf{\begin{tabular}[c]{@{}c@{}}Top-1 \\ Accuracy\\ (\%)\end{tabular}} & \textbf{\begin{tabular}[c]{@{}c@{}}Top-5 \\ Accuracy\\  (\%)\end{tabular}} \\ \hline
B0             & 5.3                                                                               & 0.39                                                                           & 224 x 224           & 77.1                                                                      & 93.3                                                                       \\ \hline
B1             & 7.8                                                                               & 0.70                                                                           & 240 x 240           & 79.1                                                                      & 94.4                                                                       \\ \hline
B2             & 9.2                                                                               & 1.00                                                                           & 260 x 260           & 80.4                                                                      & 95.0                                                                       \\ \hline
B3             & 12                                                                                & 1.80                                                                           & 300 x 300           & 81.6                                                                      & 95.7                                                                       \\ \hline
B4             & 19                                                                                & 4.20                                                                           & 380 x 380           & 82.9                                                                      & 96.4                                                                       \\ \hline
B5             & 30                                                                                & 9.90                                                                           & 456 x 456           & 83.6                                                                      & 96.7                                                                       \\ \hline
B6             & 43                                                                                & 19.0                                                                           & 528 x 528           & 84.0                                                                      & 96.8                                                                       \\ \hline
B7             & 66                                                                                & 37.0                                                                           & 600 x 600           & 84.4                                                                      & 97.0                                                                       \\ \hline
\end{tabular}
\end{table}
It uses a technique referred to as "Compound coefficient scaling" which scales the three dimensions of a network in a certain fixed fashion using a set of scaling coefficients instead of randomly scaling up the three dimensions namely depth (number of layers), width (number of channels per layer) and resolution (input image size).

It is built using mobile inverted bottleneck convolution (MBConv) blocks which was first used in MobileNetV2. The architecture of EfficientNet inlcudes many layers like convolutional layers, batch normalization, ReLU6 activations, pooling layers, and FCL but its main differentiation point is the compound coefficient scaling method which is described above. The main components of this family are described below:
\begin{enumerate}
    \item Stem: This contains the first layer of the architecture which is used to extract basic features like patterns and edges from the input image. It generally contains a convolution layer with 3x3 kernel size and a stride of 2 along with 32 output filters.
    \item Body: This consists of the base forming MBConv blocks with distinct configurations. Each MBConv block contains an Expansion layer which employs a 1x1 convolution to expand the number of features in the input feature map, a Depthwise Convolution layer which processes each feature separately, a Squeeze-and-Excitation (SE) Layer which uses a dynamic approach to adjust the significance of different features and a projection layer which, as the name suggests, projects the expanded feature map back to the original dimension using another 1x1 convolution. The main function of these blocks is to be efficient but continue maintaining high performance.
    \item Head: This consists of a last convolutional block which further refines the features received from the body block and precedes a global average pooling layer which reduces the spatial dimensions to a single value per channel (feature map). This is then succeeded by a FCL which uses these features to make the final classification decision.
\end{enumerate}

    Despite being a highly efficient and scalable model, it has certain limitations. It is difficult to fine tune as the optimization of its compound scaling factors can be tough to implement. It is also susceptible to overfitting when used with smaller datasets. Apart from this, it needs high end GPUs or TPUs for good performance.

\subsection{VGGNet}
VGGNet is another family of CNNs launched by Visual Geometry Group (VGG) at the University of Oxford in the year 2014 which is also mainly focussed on image classification. The main innovation behind it is the use of very small (3x3) convolutional filters which helps in maintaining reasonable computational complexity \cite{simonyan2014very}. Just like EfficientNet, it contains several layers like convolutional layers, max-pooling layers, ReLU activations, and FCL but the main differentiating factor is the one mentioned above and the use of 2x2 max-pooling layers throughout the network. The main components of VGGNet are described below:
\begin{enumerate}
    \item Convolutional Layers: These layers make use of 3x3 convolutional filters having a stride and padding of 1. Each convolutional layer is grouped with multiple other layers and each group is succeeded by a max-pooling layer. The number of filters increases exponentially (by a power of 2) from 64 to 512 with each layer.
    \item Max-pooling layers: These layers use 2x2 filters and have a stride of 2 to reduce the feature maps which in turn, reduces their spatial dimension (also helps in reducing computational cost) but retains important features.
    \item Fully connected layers: The convolutional and pooling layers flatten the feature maps which are then passed through FCL that perform the final classification task. The last layers then use SoftMax activation function to give the output probabalities.
    Dropout technique with a rate of 50\% is generally applied to FCL to reduce the chances overfitting. 
\end{enumerate}

There are 4 versions of VGGNet (VGG-11, VGG-13, VGG-16 and VGG-19) which are measured against FLOPS, Top-1 accuracy percent and Top-5 accuracy percent. Further details about the different versions is described in Table \ref{tab:vggnet}. VGG-16 and VGG-19 are the most widely used versions among the four.

Despite being a powerful model, it has its own set of limitations. The major disadvantage being the high computational cost on versions like VGG-16 and VGG-19 because of the huge number of training parameters. Moreover, it is slow in making predictions as it does not prioritize efficiency. This makes it less suitable for real-time applications. 
\begin{table}[!hb]
\centering
\caption{VGGNet : Versions, Parameters, Size and Accuracy}
\label{tab:vggnet}
\begin{tabular}{|c|c|c|c|c|}
\hline
\rowcolor[HTML]{EFEFEF} 
\textbf{Model} & \cellcolor[HTML]{EFEFEF}\textbf{\begin{tabular}[c]{@{}c@{}}Parameters\\ (M)\end{tabular}} & \textbf{\begin{tabular}[c]{@{}c@{}}FLOPs\\ (B)\end{tabular}} & \textbf{\begin{tabular}[c]{@{}c@{}}Top-1 \\ Accuracy\\ (\%)\end{tabular}} & \textbf{\begin{tabular}[c]{@{}c@{}}Top-5 \\ Accuracy\\  (\%)\end{tabular}} \\ \hline
VGG-11         & 132.9                                                                                     & 7.3                                                          & 69.8                                                                      & 89.8                                                                       \\ \hline
VGG-13         & 133                                                                                       & 11.32                                                        & 71.5                                                                      & 90.6                                                                       \\ \hline
VGG-16         & 138.4                                                                                     & 15.49                                                        & 71.6                                                                      & 91.0                                                                       \\ \hline
VGG-19         & 143.7                                                                                     & 19.63                                                        & 71.8                                                                      & 91.1                                                                       \\ \hline
\end{tabular}
\end{table}
\subsection{MobileNet}
MobileNet is a CNN developed by Google in the year 2017. MobileNet was specifically designed for mobile and edge devices due to their limited computing power. MobileNet heavily utilizes depthwise separable convolutions that reduce both parameters and computations without losing significant performance \cite{howard2017mobilenets}.

MobileNet is built with two major steps, which are depthwise and pointwise convolutions (1x1 convolution). This allows the computer to apply one filter per feature, capture important spatial representations, and create, rich representations for further use, leading to increased efficiency relative to traditional CNNs. Other notable features are: batch normalization (which helps to stabilize the training process), ReLU activation (to create non-linearity in the output), and stride convolutions (which help to downscale the image). The network is finally completed by global average pooling, a FCL and SoftMax classifier to perform the prediction tasks.

MobileNet has evolved into multiple versions, each with improving efficiency and accuracy:
\begin{itemize}
\item MobileNet V1: It is the original version that introduced depthwise separable convolutions.
\item MobileNet V2: It introduced inverted residual blocks and linear bottlenecks which allowed for better information flow and efficiency.
\item MobileNet V3: It made use of Neural Architecture Search (NAS) to optimize the network and incorporated SE modules to enhance feature selection. Also introduced the Hard Swish activation function to improve mobile performance.
\end{itemize}
Further details about the different versions of MobileNet is given in Table \ref{tab:mobilenet}.

MobileNet is widely used in various applications such as including image classification, object detection, and facial recognition. It has usage in medical imaging such as COVID-19 and skin cancer detection. In agriculture, it is used for identifying plant diseases. While in industries, it is used to detect quality control such as welding defect detection. Its key advantage is that it requires low computational cost, has a small model size, and has real-time processing capabilities, which makes it ideal for deployment on mobile and edge devices.

\begin{table}[]
\centering
\caption{MobileNet : Versions, Parameters, Size and Accuracy}
\label{tab:mobilenet}
\begin{tabular}{|c|c|c|c|c|}
\hline
\rowcolor[HTML]{EFEFEF} 
\textbf{Model}                                               & \cellcolor[HTML]{EFEFEF}\textbf{\begin{tabular}[c]{@{}c@{}}Parameters\\ (M)\end{tabular}} & \textbf{\begin{tabular}[c]{@{}c@{}}FLOPs\\ (B)\end{tabular}} & \textbf{\begin{tabular}[c]{@{}c@{}}Top-1 \\ Accuracy\\ (\%)\end{tabular}} & \textbf{\begin{tabular}[c]{@{}c@{}}Top-5 \\ Accuracy\\  (\%)\end{tabular}} \\ \hline
\begin{tabular}[c]{@{}c@{}}MobileNet\\ V1\end{tabular}       & 4.2                                                                                       & 0.57                                                         & 70.6                                                                      & 89.5                                                                       \\ \hline
\begin{tabular}[c]{@{}c@{}}MobileNet\\ V2\end{tabular}       & 3.4                                                                                       & 0.30                                                         & 71.8                                                                      & 91.0                                                                       \\ \hline
\begin{tabular}[c]{@{}c@{}}MobileNet\\ V3 Small\end{tabular} & 2.5                                                                                       & 0.06                                                         & 67.4                                                                      & 87.4                                                                       \\ \hline
\begin{tabular}[c]{@{}c@{}}MobileNet\\ V3 Large\end{tabular} & 5.4                                                                                       & 0.22                                                         & 75.2                                                                      & 92.2                                                                       \\ \hline
\end{tabular}
\end{table}

While MobileNet is efficient, it may not reach the same accuracy levels as larger models, particularly when it comes to handling more complex tasks.
\subsection{ResNet}
Residual Network (ResNet)  is a deep CNN architecture that was first introduced by Microsoft back in 2015. It makes use of skip connections that are also known as residual connections to allow gradients to bypass certain layers. By doing this it is easily able to overcome the issue of vanishing gradients, which makes optimization difficult as the number of layers increases. This makes training very deep networks easier and ensures that they can be trained effectively without degradation of performace\cite{he2016deep}.

ResNet’s architecture includes convolutional layers, batch normalization, ReLU activations, pooling layers, and FCL. However, its most defining feature is the residual block that it contains. These blocks make use of identity shortcuts to bypass certain layers, which ensures smooth gradient flow and preserves important information. Unlike most traditional CNNs, that can lose accuracy as the number of layers increase, ResNet allows the layers to learn residual mappings, which makes deep networks much more efficient. The early layers capture basic spatial features, while the deeper layers extract more complex patterns. By improving the training speed and accuracy, ResNet has become a very powerful architecture for DL applications.

ResNet has evolved into multiple versions, each of which is distinguished by the number of layers it contains:
\begin{itemize}
\item ResNet-18 and ResNet-34: They are shallower versions that offer a pretty good balance between computational efficiency and accuracy.
\item ResNet-50: Its a widely used variant that maintains a strong performance while incorporating bottleneck layers for better optimizations.
\item ResNet-101 and ResNet-152: They are deeper models that are designed for large-scale image recognition tasks. While they require more computational resources, they provide enhanced accuracy.
\end{itemize}
Further details about the different versions of ResNet is given in Table \ref{tab:resnet}.
\begin{table}[]
\centering
\caption{ResNet : Versions, Parameters, Size and Accuracy}
\label{tab:resnet}
\begin{tabular}{|c|c|c|c|c|}
\hline
\rowcolor[HTML]{EFEFEF} 
\textbf{Model}                   & \cellcolor[HTML]{EFEFEF}\textbf{\begin{tabular}[c]{@{}c@{}}Parameters\\ (M)\end{tabular}} & \textbf{\begin{tabular}[c]{@{}c@{}}FLOPs\\ (B)\end{tabular}} & \textbf{\begin{tabular}[c]{@{}c@{}}Top-1 \\ Accuracy\\ (\%)\end{tabular}} & \textbf{\begin{tabular}[c]{@{}c@{}}Top-5 \\ Accuracy\\  (\%)\end{tabular}} \\ \hline
ResNet-18                        & 11.7                                                                                      & 1.8                                                          & 69.6                                                                      & 89.1                                                                       \\ \hline
ResNet-34                        & 21.8                                                                                      & 3.6                                                          & 73.2                                                                      & 91.3                                                                       \\ \hline
ResNet-50                        & 25.6                                                                                      & 4.1                                                          & 76.0                                                                      & 93.0                                                                       \\ \hline
ResNet-101                       & 44.5                                                                                      & 7.9                                                          & 77.4                                                                      & 93.7                                                                       \\ \hline
\multicolumn{1}{|l|}{ResNet-152} & $\sim$60                                                                                  & $\sim$11                                                     & $\sim$77.8                                                                & $\sim$93.8                                                                 \\ \hline
\end{tabular}
\end{table}
ResNet is widely used for image classification, object detection, segmentation, and facial recognition applications. It serves as a ground model for advanced DL architectures such as the Faster R-CNN and the Mask R-CNN, which are usually used in fields such as medical imaging, autonomous driving, and industrial defect detection. 

Despite its many advantages, ResNet does have a certain limitations. Its deeper variants, like ResNet-101 and ResNet-152, require a substantial amount of computational power and memory, which make them less ideal for usage in real-time applications on low-resource devices. Additionally, while the residual connections help us enhance training, they also introduce computational overhead that slows down the inference speed.

\subsection{Optimizers}

Optimizers are essentially algorithms that are used to adjust the model's parameters to minimize the loss function during its training. They determine how the model learns from the given data by controlling the step size and the direction of parameter updates. Popular optimizers include Adaptive Moment Estimation (Adam) Optimizer and SGD Optimizer.
\subsubsection{Adam Optimizer}
Adam optimizer is a new and advanced design introduced in 2015 which tries to optimize the gradient-based learning process by estimating an individual learning rate for every parameter. Adam takes the advantages of two other derivatives AdaGrad which captures and holds onto learning rates that improves performance on sparse gradient problems and RMSProp which adjusts learning rate for each parameter based on the magnitude of the recent gradient.

The Adam optimizer is especially useful for training deep neural networks in the presence of noisy data, non-stationary objectives, and sparse gradient data. Its flexibility is the reason why it is frequently used in many DL fields, including computer vision, NLP, and RL, where older approaches had difficulty achieving efficient convergence.

The mathematical formulation of Adam involves computing adaptive learning rates for each parameter by maintaining estimates of the first and second moments of the gradients. The general formulae that are applied are as follows\cite{kingma2014adam}:
 
\begin{enumerate}  
\item Update the biased first-moment estimate:  
\[
p_t = \gamma_1 p_{t-1} + (1 - \gamma_1) q_t
\]  
where \( p_t \) represents the exponentially weighted moving average of the gradients, and \( \gamma_1 \) is the decay rate.  

\item Update the biased second raw moment estimate:  
\[
r_t = \gamma_2 r_{t-1} + (1 - \gamma_2) q_t^2
\]  
where \( r_t \) tracks the moving average of squared gradients, controlled by the decay factor \( \gamma_2 \).  

\item Compute bias-corrected moment estimates:  
\[
\tilde{p}_t = \frac{p_t}{1 - \gamma_1^t}, \quad \tilde{r}_t = \frac{r_t}{1 - \gamma_2^t}
\]  
where \( \tilde{p}_t \) and \( \tilde{r}_t \) are bias-corrected estimates to ensure unbiased gradient estimates in early iterations.  

\item Update the parameters:  
\[
\psi_t = \psi_{t-1} - \frac{\eta \tilde{p}_t}{\sqrt{\tilde{r}_t} + \delta}
\]  
where \( \psi_t \) represents the updated model parameters, \( \eta \) is the learning rate, and \( \delta \) is a small constant for stability.  
\end{enumerate}

Adam is a popular choice in DL optimizer because of its various advantages. While it is highly computationally efficient and has low memory consumption, which is beneficial for large models, its primary advantage is its invariant to diagonal gradient rescaling and therefore is stable with respect to different magnitudes of parameter values. Furthermore, it is effective in dealing with DL objectives, which have non-stationary and noisy gradients. Historically, Adam did well with tasks with frameworks with sparse gradients unlike other traditional optimizers. Another benefit is that Adam did not need many hyperparameter tunings as default settings performed decently in most applications.

For all its benefits, Adam suffers from some problems. It had issues performing a form of inference, especially in specific architectures of deep networks. While it is skilled in efficiently adapting the learning rates, it is less effective in other use cases such as in computer vision. Additionally, the way it applied weight decay contradicts most other methods and lead to implementation of other methods such as AdamW. Along with this, Adam needed more memory, which was a concern for certain, resource-poor systems. Lastly, while it was adaptive, the fine-tuning of the learning rate was problematic.
\subsubsection{SGD Optimizer}
SGD optimizer is one of the most fundamental optimization algorithm that is applied in ML and DL. It was first introduced back in 1951.  It is commonly used in fields like computer vision, speech recognition, and large-scale data analysis. It shines in scenarios where the computational efficiency is essential. 

One of the key features of SGD is that it uses small batches of data to update weights instead of the entire datasets which greatly speeds up computations and training time, especially while working with large datasets.\cite{robbins1951stochastic}. 
The update rule for SGD is given by the following mathematical formula:
\begin{equation}
\psi_{t+1} = \psi_t - \eta \nabla J(\psi_t)
\end{equation}

where:
\begin{itemize}
    \item \( \psi_t \) represents the model parameters at step \( t \),
    \item \( \eta \) is the learning rate,
    \item \( \nabla J(\psi_t) \) is the gradient of the objective function at \( \psi_t \).
\end{itemize}
The stochasticity present in SGD provides desirable noise that helps the models escape sharp local minima, which leads to improved generalization performance. SGD is also memory-efficient and easily scalable to more complex applications and offers an easy implementation with a very low number of hyperparameters, which makes this method very favorable. These advantages are why it is commonly used in DL applications, particularly in training CNNs for image classification and object detection.
In many cases, SGD is found to outperform adaptive methods in terms of convergence and final model performance.
Despite its widespread use, SGD does have a few limitations. The variance that is present in updates due to the small batch sampling results in a lot of instability being created during the training of the dataset. This led to the development of the variant called the  Mini-Batch SGD, where, instead of using a single data point, the parameters are updated using a small batch of data points\cite{bottou2010large}.
SGD is particularly sensitive to the selection of the learning rate. An improper learning rate choice often leads to slow convergence or divergence. Also, since SGD does not adjust the learning rates per parameter, it is prone to getting trapped in local minima(s) when faced with a complex optimization situation resulting in a significantly slower convergence.
\begin{figure*}[!h]
    \centering
    \includegraphics[width=0.7\linewidth]{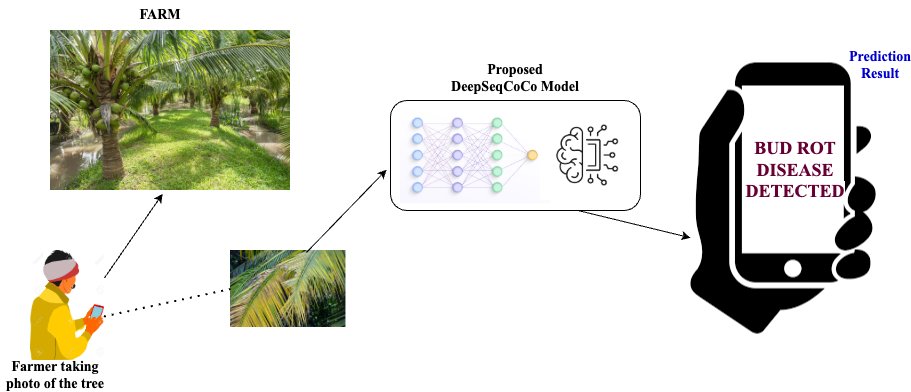}
    \caption{Illustration of the use-case}
    \label{fig:usecase}
\end{figure*}

\section{Proposed Methodology}

\begin{figure*}[!h]
    \centering
    \includegraphics[width=0.7\linewidth]{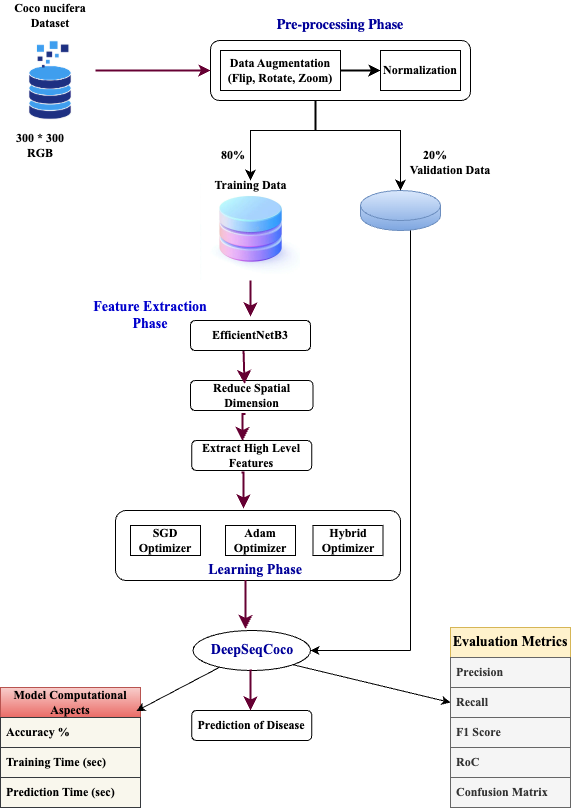}
    \caption{Proposed DeepSeqCoCo Deep Learning Framework}
    \label{fig:DeepSeqCoCo}
\end{figure*}
The agriculturalists have an expectation on the yield of their cultivation when they first invest in their farm land. But at a later stage, due to environmental conditions, pests, and insects, the yield is reduced sometimes. In many cases, the farmers are negligent that they do not notice if the coconut trees are affected by any diseases at an early stage so that they can take some measures to save the tree and therefore maintain their yield. Even if they notice some changes in the trees’ growth they are unaware that the tree is affected by disease. When such differences are noticed at an early stage, some measures can be taken to cure them so that the trees will become healthy. Our proposed methodology focuses on providing a technical assistance to farmers, where they can just take a picture of the tree using their smart phone and request for support. Our model would assist them by predicting the type of disease by which the coconut tree is affected as illustrated in Figure \ref{fig:usecase}. Using this, the farmers can take steps to cure the disease using some prescribed pesticides or insecticides for the predicted disease. This can help improving their yield and increase their profit.

The proposed DeepSeqCoco Model has totally seven different layers each comprising of various levels performing multiple functionalities. The layers are Input layer, Data Augmentation layer, Normalization layer, Feature extractor layer, Average pooling layer, Dense layer and Prediction layer. Figure \ref{fig:DeepSeqCoCo} illustrates the architecture of the model.
The Input layer in the model is the initial layer which takes a RGB image as input of size 300 x 300. This layer just acts as a place holder for the input image before any processing is performed. Every neuron in the first layer is considered to be equivalent to a pixel in the image. Hence, we may consider that there are around 2,70,000 neurons. Once the dataset is divided into training and validation set based on the 80-20 rule, the training set images are passed through the data augmentation layer, where the images are augmented by first randomly flipping the images horizontally, then rotating them by ±20\% and finally zooming the images by ±20\%. For increasing the robustness of the model, this layer is added. This also helps in preventing over-fitting issues while training. The normalization layer helps in scaling each pixel value from (0-255) to (0-1). Then comes the feature extraction layer which is based on the pre-trained EfficientNet B3 model. The structure of the feature extractor model is shown in the Table \ref{tab:feature}.  \begin{table}[!h]
\centering
\caption{Structure of Feature Extractor}
\label{tab:feature}
\begin{tabular}{|ccccc|}
\hline
\multicolumn{5}{|c|}{\cellcolor[HTML]{EFEFEF}\textbf{Convolution and Pooling Layer (3 x 3, 40 filters, stride=2)}}                                                                                                                                                                             \\ \hline
\multicolumn{5}{|c|}{\cellcolor[HTML]{EFEFEF}\textbf{Mobile Inverted Bottleneck Convolution}}                                                                                                                                                                                                                           \\ \hline
\multicolumn{1}{|c|}{\textbf{Stage}} & \multicolumn{1}{c|}{\textbf{Operator}} & \multicolumn{1}{c|}{\textbf{\begin{tabular}[c]{@{}c@{}}No.of \\ Output \\ Channels\end{tabular}}} & \multicolumn{1}{c|}{\textbf{\begin{tabular}[c]{@{}c@{}}No. \\ of\\ Layers\end{tabular}}} & \textbf{Stride} \\ \hline
\multicolumn{1}{|c|}{1}              & \multicolumn{1}{c|}{MBConv1}           & \multicolumn{1}{c|}{24}                                                                           & \multicolumn{1}{c|}{3}                                                                   & 1               \\ \hline
\multicolumn{1}{|c|}{2}              & \multicolumn{1}{c|}{MBConv6}           & \multicolumn{1}{c|}{32}                                                                           & \multicolumn{1}{c|}{3}                                                                   & 2               \\ \hline
\multicolumn{1}{|c|}{3}              & \multicolumn{1}{c|}{MBConv6}           & \multicolumn{1}{c|}{48}                                                                           & \multicolumn{1}{c|}{3}                                                                   & 2               \\ \hline
\multicolumn{1}{|c|}{4}              & \multicolumn{1}{c|}{MBConv6}           & \multicolumn{1}{c|}{96}                                                                           & \multicolumn{1}{c|}{5}                                                                   & 2               \\ \hline
\multicolumn{1}{|c|}{5}              & \multicolumn{1}{c|}{MBConv6}           & \multicolumn{1}{c|}{136}                                                                          & \multicolumn{1}{c|}{5}                                                                   & 1               \\ \hline
\multicolumn{1}{|c|}{6}              & \multicolumn{1}{c|}{MBConv6}           & \multicolumn{1}{c|}{232}                                                                          & \multicolumn{1}{c|}{5}                                                                   & 2               \\ \hline
\multicolumn{1}{|c|}{7}              & \multicolumn{1}{c|}{MBConv6}           & \multicolumn{1}{c|}{384}                                                                          & \multicolumn{1}{c|}{3}                                                                   & 1               \\ \hline
\multicolumn{5}{|c|}{\cellcolor[HTML]{EFEFEF}\textbf{Convolution (1 x 1)}}                                                                                                                                                                                                                     \\ \hline
\end{tabular}
\end{table} 

The very first layer in this feature extractor is the convolution layer which consists of 40 filters of size 3 x 3, and the filter is configured to move 2 pixels at a time. This is the top layer of the feature extraction and captures the low-level features from the image. As the stride is fixed as 2, the image resolution is reduced from 300 x 300 to 150 x 150. The output of this layer is a feature map of size 150 x 150. Totally 40 feature maps are generated at the end of this convolution operation.

The layer succeeding is the MBConv block having 24 filters. This block consists of seven sub-blocks. The first block represented by MBConv1 is an optimized depth-wise convolution which consists of 3 layers, namely Depth-wise Convolution (DWConv), Point-wise Convolution, and SE. DWConv applies a 3 x 3 convolution to every channel separately and tries to extract the spatial features. The next layer is the Point-wise Convolution which is of shape 1 x 1. This layer combines the features from different channels and compresses or expands the number of feature maps as per the requirement. SE uses a channel-wise attention mechanism for shifting the focus towards learning important features of the image. The final output is of shape 150 x 150 x 24 as the stride used was 1.

The next set of layers are MBConv6, which means that they have an expansion factor of 6. As depicted in Table \ref{tab:feature}, the number of filters increases while moving deep, for capturing more complex and abstract features. When the count of filters increases, the spatial size decreases. The final convolution layer has a total of 1536 filters, and the output shape of the feature map is 10 x 10 x 1536. The final feature vector of size 1 x 1 x 1536 is obtained via the average pooling layer.

The Stem layer, which is the initial convolution and pooling layer, has 40 filters with a 3 x 3 kernel size, and the filter moves at a stride value of 2. This reduces the spatial dimension by 50\%. This layer helps in extracting the low-level features like edges and textures. The MBConv blocks are used for improving efficiency.

There is a final 1 x 1 convolutional layer that increases the channels to 1536. The deep MB-Conv layers help in capturing abstract as well as complex features. The average pooling layer converts the feature maps of dimension 10 x 10 x 1536 into a feature vector of size 1 x 1536. The feature vector obtained is then passed over to the fully connected dense layer. This layer has 5 neurons since there are a total of 5 classes of diseases. For obtaining the probability distribution, the softmax activation function is utilized. Finally, the output layer provides the disease prediction. It is a vector having 5 values. Each value corresponds to the probability of the class. The class with the highest probability value is considered to be the final prediction.

\section{Results and Experimental Analysis}
This section starts with specifying the details of the data set used in the validation of the proposed framework. This is followed by the 
experimental setup and finally the results obtained and comparison with related works.
\subsection{Dataset Description}
The dataset used for the model was taken from Mendeley and is titled "Coconut Tree Disease Dataset". It is the first open access dataset on coconut tree diseases and was published in the year 2023.

Patil \textit{et al.} collected images of coconut trees from a region named \textit{Kendur} in Pune district of Maharashtra, India. The region has vast areas of coconut plantations and varying environmental conditions which was the primary reason of choosing it. The images were captured using a smartphone camera in the months of April to July during daylight to ensure proper lighting conditions. In total, 5798 images were captured and preprocessed using IrfanView software to establish uniformity and carefully highlight disease features \cite{thite2023coconut} \cite{patil2023coconut}. More details about the images are given in Table \ref{tab:data}.
\begin{table}[]
\centering
\caption{Details of Dataset used in Experimentation}
\label{tab:data}
\begin{tabular}{|
>{\columncolor[HTML]{EFEFEF}}l |l|}
\hline
\multicolumn{1}{|c|}{\cellcolor[HTML]{EFEFEF}\textbf{Attribute}}           & \multicolumn{1}{c|}{\cellcolor[HTML]{EFEFEF}\textbf{Details}}                                                  \\ \hline
\textbf{Dataset Name}                                                      & Coconut Tree Disease Dataset                                                                                   \\ \hline
\textbf{Total Images}                                                      & 5,798                                                                                                          \\ \hline
\textbf{Image Resolution}                                                  & 72 dpi                                                                                                         \\ \hline
\textbf{Image Dimension}                                                   & 768 x 1,024 pixels                                                                                             \\ \hline
\textbf{Capture Device}                                                    & \begin{tabular}[c]{@{}l@{}}Samsung Galaxy F23 5G (50 Megapixel \\ Sony IMX582 Sensor)\end{tabular}             \\ \hline
\textbf{\begin{tabular}[c]{@{}l@{}}Data Acquired \\ Location\end{tabular}} & \begin{tabular}[c]{@{}l@{}}Kendur, Taluka - Shirur, Pune, Maharastra,\\ India\end{tabular}                     \\ \hline
\textbf{\begin{tabular}[c]{@{}l@{}}Preprocessing \\ Steps\end{tabular}}    & \begin{tabular}[c]{@{}l@{}}Resizing, Cropping, Contrast adjustment \\ using IrfanView Software\end{tabular}    \\ \hline
\textbf{Labelling Process}                                                 & \begin{tabular}[c]{@{}l@{}}Expert validation by plant pathologists and\\ agricultural specialists\end{tabular} \\ \hline
\end{tabular}
\end{table}
The images belonged to 5 classes of diseases namely,
\begin{enumerate}
    \item Bud Root Dropping
    \item Bud Rot
    \item Gray Leaf Spot
    \item Leaf Rot
    \item Stem Bleeding
\end{enumerate}
Figure \ref{Fig:Classes} below shows some sample pictures of trees affected by these diseases, while Table \ref{tab:disease} explains the key characteristics and impact of the 5 diseases mentioned above. It also mentions the number of images belonging to each class in the dataset.

\begin{figure}[htbp] 
    \centering 
    \begin{tabular}{ccc}
        \includegraphics[width=25mm]{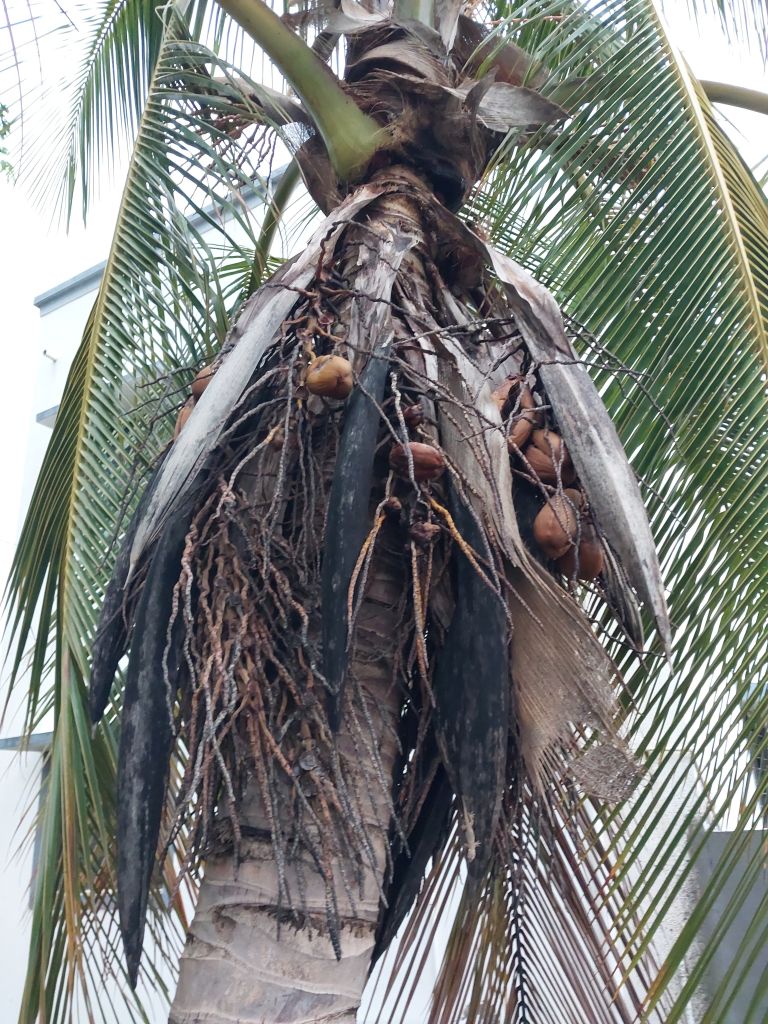} &
        \includegraphics[width=25mm]{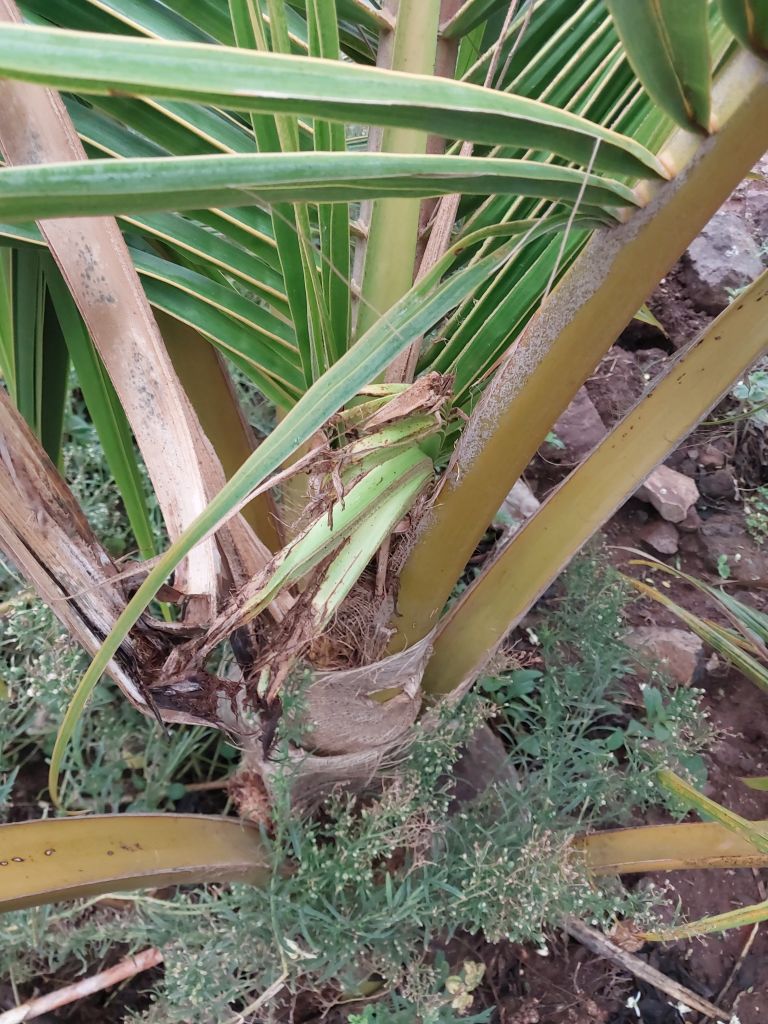} &
        \includegraphics[width=25mm]{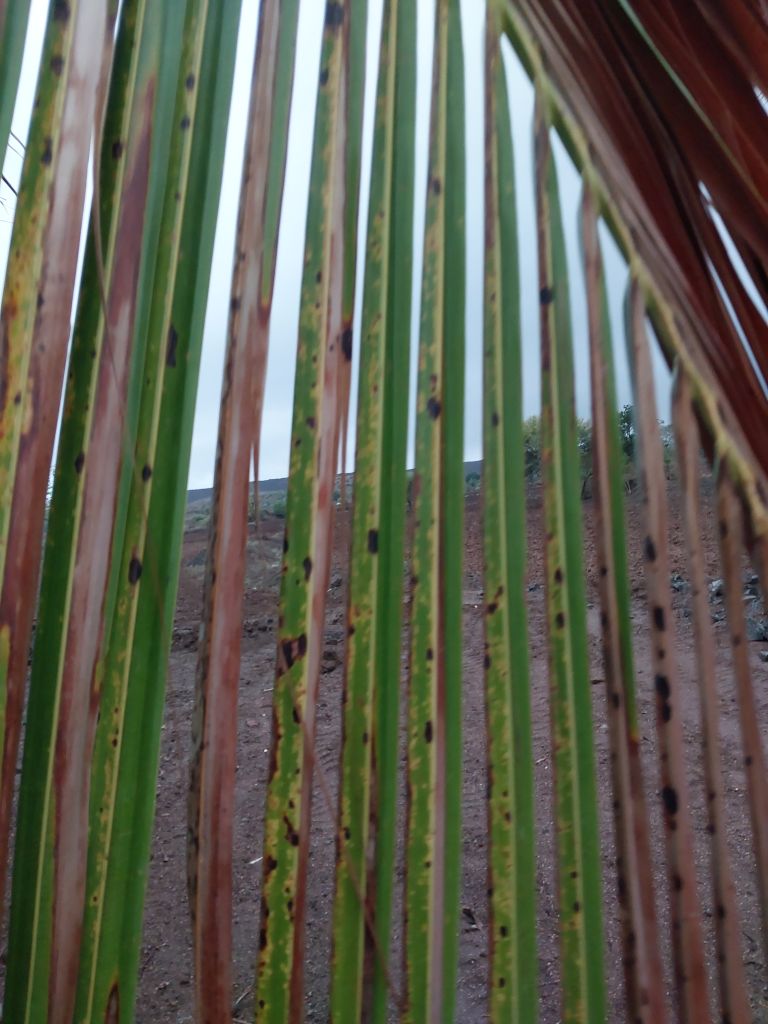} \\
        (a) & (b) & (c) \\
        \includegraphics[width=25mm]{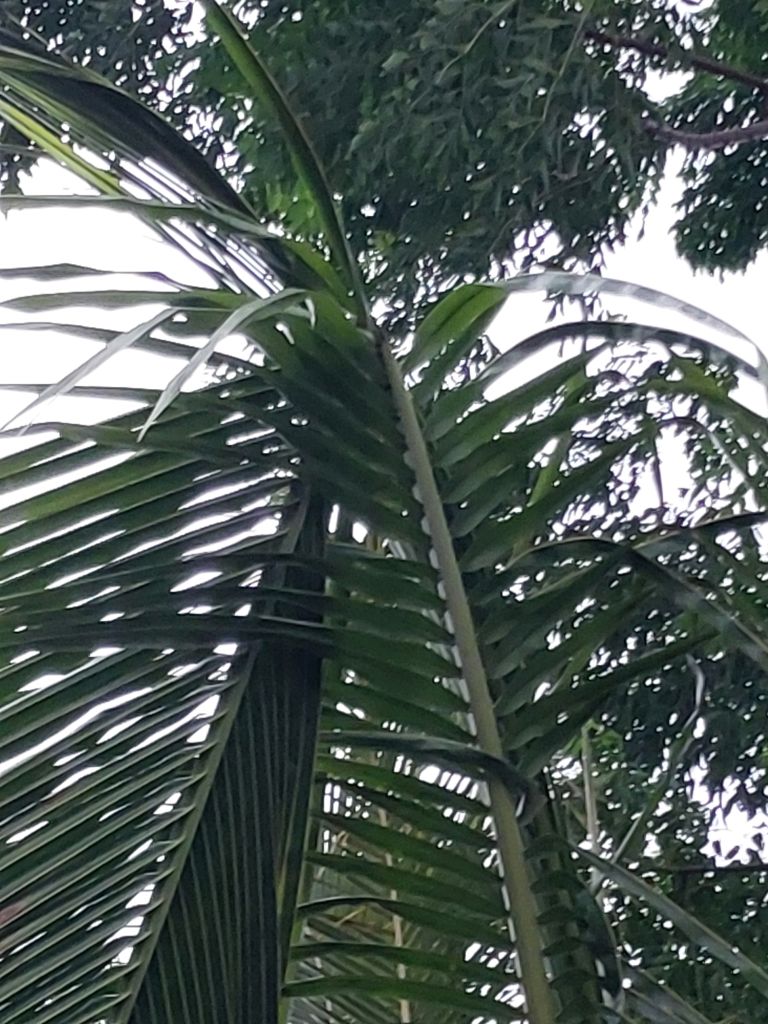} &
        \includegraphics[width=25mm]{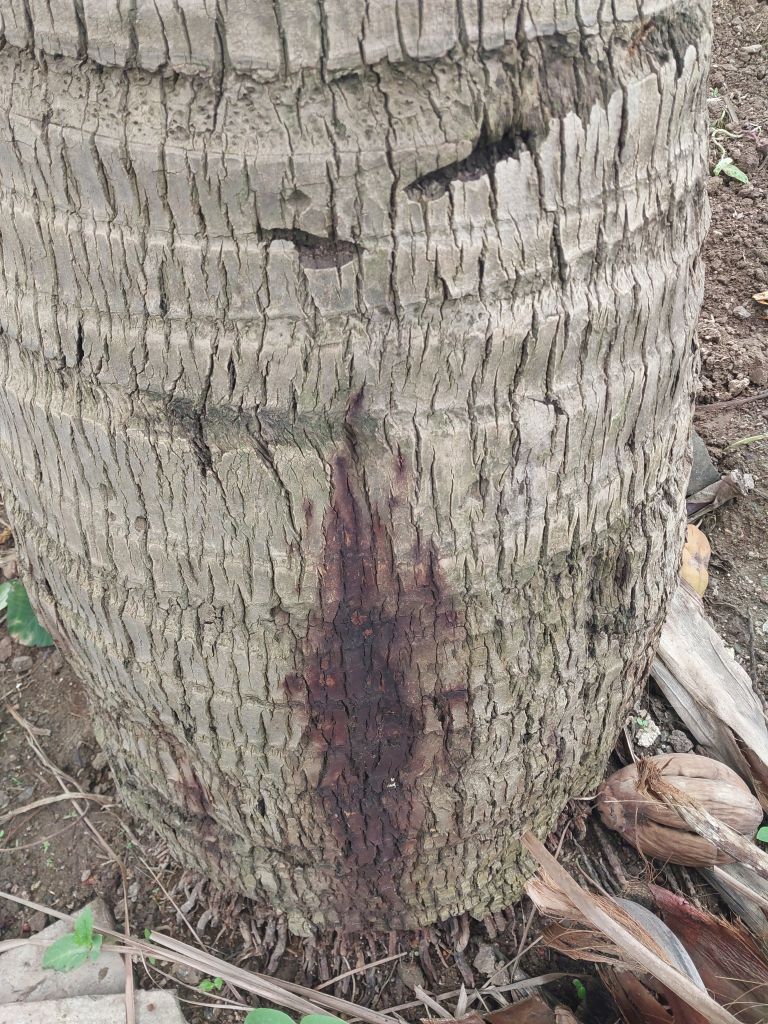} &
        \\ 
        (d) & (e) & 
    \end{tabular}
    \caption{Target Classes: (a) Bud Root Dropping (b) Bud Rot (c) Gray Leaf Spot (d) Leaf Rot (e) Stem Bleeding} 
    \label{Fig:Classes}
\end{figure}

\begin{table*}[!h]
\centering
\caption{Coconut Tree: Types of Diseases and Characteristics }
\label{tab:disease}
\begin{tabular}{|l|l|l|l|}
\hline
\rowcolor[HTML]{EFEFEF} 
\multicolumn{1}{|c|}{\cellcolor[HTML]{EFEFEF}\textbf{Disease Name}} & \multicolumn{1}{c|}{\cellcolor[HTML]{EFEFEF}\textbf{\begin{tabular}[c]{@{}c@{}}No.of \\ Images\end{tabular}}} & \multicolumn{1}{c|}{\cellcolor[HTML]{EFEFEF}\textbf{Key Characteristics}}                                                                                          & \multicolumn{1}{c|}{\cellcolor[HTML]{EFEFEF}\textbf{Impact}}                                              \\ \hline
\begin{tabular}[c]{@{}l@{}}Gray Leaf \\ Spot\end{tabular}           & 2.135                                                                                                         & \begin{tabular}[c]{@{}l@{}}* Fungal Disease\\ * Creates grayish spots on leaves\\ * Usually starts at leaf margins\end{tabular}                                    & \begin{tabular}[c]{@{}l@{}}Spreads quickly and \\ has the potential to cause \\excess leaf loss\end{tabular}               \\ \hline
Leaf Rot                                                            & 1,673                                                                                                         & \begin{tabular}[c]{@{}l@{}}* Progressive decay of leaf tissue\\ * Often shows brown/black discoloration\\ * Can affect multiple leaves simultaneously\end{tabular} & \begin{tabular}[c]{@{}l@{}}Affects photosynthesis \\ ability of the tree\end{tabular}                     \\ \hline
Stem Bleeding                                                       & 1,006                                                                                                         & \begin{tabular}[c]{@{}l@{}}* Characteristic reddish-brown fluid oozing  \\    from trunk\\ * Often appears as dark patches on stem\end{tabular}                    & \begin{tabular}[c]{@{}l@{}}Weakens tree's structural \\ integrity\end{tabular}                            \\ \hline
\begin{tabular}[c]{@{}l@{}}Bud Root \\ Dropping\end{tabular}        & 514                                                                                                           & \begin{tabular}[c]{@{}l@{}}* Causes issues in early growth stages\\ * Root system deterioration\\ * Visible wilting of buds\end{tabular}                                    & \begin{tabular}[c]{@{}l@{}}Can result in decreased\\ output and impaired\\ development\end{tabular}               \\ \hline
Bud Rot                                                             & 470                                                                                                           & \begin{tabular}[c]{@{}l@{}}* Fungal infection of the bud region\\ * Can cause crown rot\\ * Often fatal if untreated\end{tabular}                                  & \begin{tabular}[c]{@{}l@{}}Can quickly destroy the entire\\ plantation if not detected early\end{tabular} \\ \hline
\end{tabular}
\end{table*}
\subsection{Experimental Setup}
In order to make sure that an efficient and standardized evaluation of all the models can take place, Google Colab TPU v2-8 was used for training and validation. The dataset was preprocessed in several steps and fed to the models after a well-defined pipeline.\\ \\
\begin{enumerate}
    \item \textbf{Computing Environment}
    
    The experiments were conducted on Google Colab using a v2-8 TPU for faster model training and validation. The reason behind choosing TPUs was their ability to handle complex DL tasks efficiently. The additional details are as follows:
    \begin{itemize}
        \item Software Frameworks: TensorFlow, TensorFlow Hub, scikit-learn, Matplotlib, Seaborn, and PIL
        \item Python Version: 3.11.11
        \item TensorFlow Version: 2.15.0
        \item TensorFlow Hub Version: 0.16.1
    \end{itemize}
    \item \textbf{Data Preprocessing}

    In order to make sure of the uniformity across all input images, the dataset was preprocessed using the following techniques:
    \begin{itemize}
        \item Resizing: All images were resized to 300 × 300 pixels while maintaining the three color channels (RGB), resulting in an input shape of 300 × 300 × 3.
        \item Data Augmentation: To make sure that the model learns every aspect of the image properly, augmentation strategies like random rotation (both horizontal and vertical), random zooming, and random flipping, were carried out on the images while maintaining the image shape of 300 × 300 × 3.
        \item Normalization: Pixel values were scaled to the range [0,1] range using a rescaling layer to improve stability during training.
        \item Dataset Splitting:
        \begin{itemize}
            \item 80\% of the dataset was used for training i.e, out of 5858 images, 4687 images were used for training to ensure that there are enough images for the model to learn properly.
            \item The remaining 20\% of the dataset was used for validation which accounts to 1171 images out of the total 5858 images.
        \end{itemize}
        \item Performance Optimization Techniques:
        \begin{itemize}
            \item Pre-fetching was applied using the \textbf{AUTOTUNE} function from the Tensorflow library to ensure efficient data loading.
            \item Caching was employed to prevent repetitive computations and improve training speed by reducing redundancy and ensuring optimal resource utilization.

        \end{itemize}
    \end{itemize}

    \item \textbf{Base Model Selection}
    
    The proposed model was built using \textbf{EfficientNet-B3} as the feature extractor model. It was imported using Tensorflow Hub.
    \item \textbf{Training Procedure}
    \begin{itemize}
        \item The images were sent to the image pipeline offered by Tensorflow for batching and augmentation.
        \item Batch size was fixed at 32 to balance computational efficiency and convergence stability.
        \item \textbf{Sparse Categorical Crossentropy} was used as the loss function because it is suitable for multi-class classification.
    \end{itemize}
    \item \textbf{Experimental Considerations}
    \begin{itemize}
        \item A fixed random seed was used for dataset splitting and shuffling in order to ensure reproducibility.
        \item Multiple evaluation criteria like accuracy, loss, F1-score and confusion matrix were used to assess model performance. This helps in ensuring a thorough evaluation of the model and remove any ambiguity related to it.
    \end{itemize}
\end{enumerate}
\subsection{Results and Discussion}
The proposed DeepSeqCoco model was experimented with using Mendeley data, which is discussed in the previous subsection. The experiment was carried out in four distinct ways. The first two experiments were performed by training the dataset using the proposed model using the optimizers individually, and the latter two were by hybridizing the optimizers. The results obtained are discussed in detail below:

\subsubsection{Learning with Other Two Versions of EfficientNet}
Initially, for deciding on the version to be used as the feature extraction layer, we performed experimental analysis using all the three variations of EfficientNet that is B1, B2 and B3. Based on the initial level of results, we came to the decision to utilize EfficientNetB3. When the EfficientNetB1 and EfficientNetB2 were used the number of parameters to be trained were 6405 and 7045 respectively as tabulated in the Table \ref{tab:versions}.

\begin{table}[]
\centering
\caption{Model Summary: EfficientNet B1 and B2}
\label{tab:versions}
\begin{tabular}{lllll}
\hline
\multicolumn{1}{c}{\multirow{2}{*}{\textbf{\begin{tabular}[c]{@{}c@{}}Layer \\ Type\end{tabular}}}} & \multicolumn{2}{c}{\textbf{EfficientNetB1}}                                                                                    & \multicolumn{2}{c}{\textbf{EfficientNetB2}}                                                                                    \\ \cline{2-5} 
\multicolumn{1}{c}{}                                                                                & \multicolumn{1}{c}{\textbf{\begin{tabular}[c]{@{}c@{}}Output \\ Shape\end{tabular}}} & \multicolumn{1}{c}{\textbf{Parameters}} & \multicolumn{1}{c}{\textbf{\begin{tabular}[c]{@{}c@{}}Output \\ Shape\end{tabular}}} & \multicolumn{1}{c}{\textbf{Parameters}} \\ \hline
keras                                                                                               & (None, 1280)                                                                         & 6575232                                 & (None, 1408)                                                                         & 7768562                                 \\ \hline
dense                                                                                               & (None, 5)                                                                            & 6405                                    & (None, 5)                                                                            & 7045                                    \\ \hline
\end{tabular}
\end{table}

\begin{figure}[!h]
    \centering
    \includegraphics[width=0.8\linewidth]{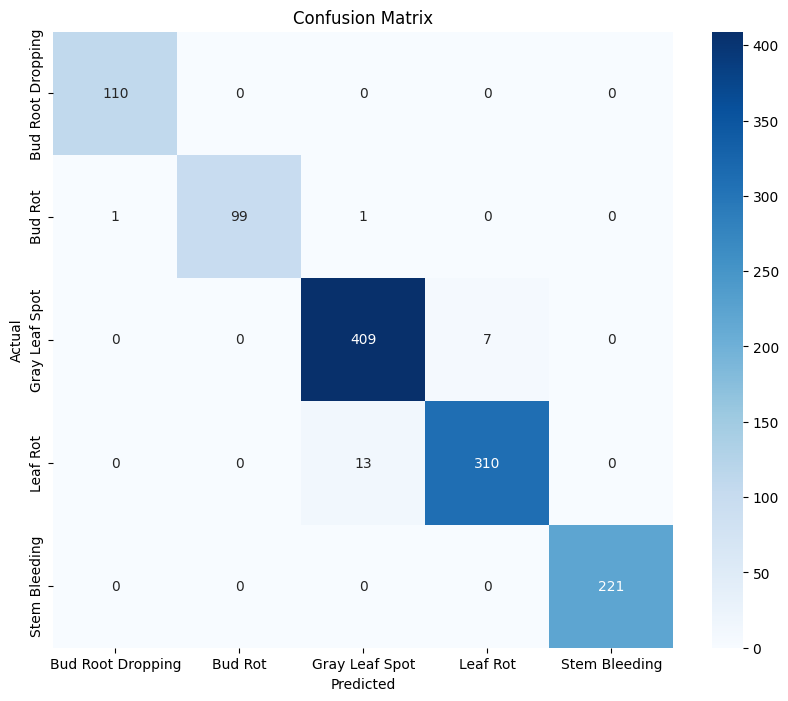}
    \caption{Confusion Matrix: EfficientNetB1 }
    \label{fig:confusion-b1}
\end{figure}

The accuracy and loss achieved when using EfficientNetB1 were 90.7\% and 31.6\%, respectively, for the training set whereas, the validation accuracy and loss were 98.1\% and 10.1\%, respectively. The time taken to learn the parameters in the training was 1094.25 seconds per epoch. The confusion matrix obtained using the testing data is shown in Figure \ref{fig:confusion-b1}.

The training accuracy and loss when using EfficientNetB2 were 91.2\% and 32.7\%, respectively, whereas, the validation accuracy and loss were 98\% and 10.4\%, respectively. The time taken for training was 1138.34 seconds per epoch. The confusion matrix obtained using the validation set is shown in Figure \ref{fig:confusion-b2} below. 
\begin{figure}[!h]
    \centering
    \includegraphics[width=0.8\linewidth]{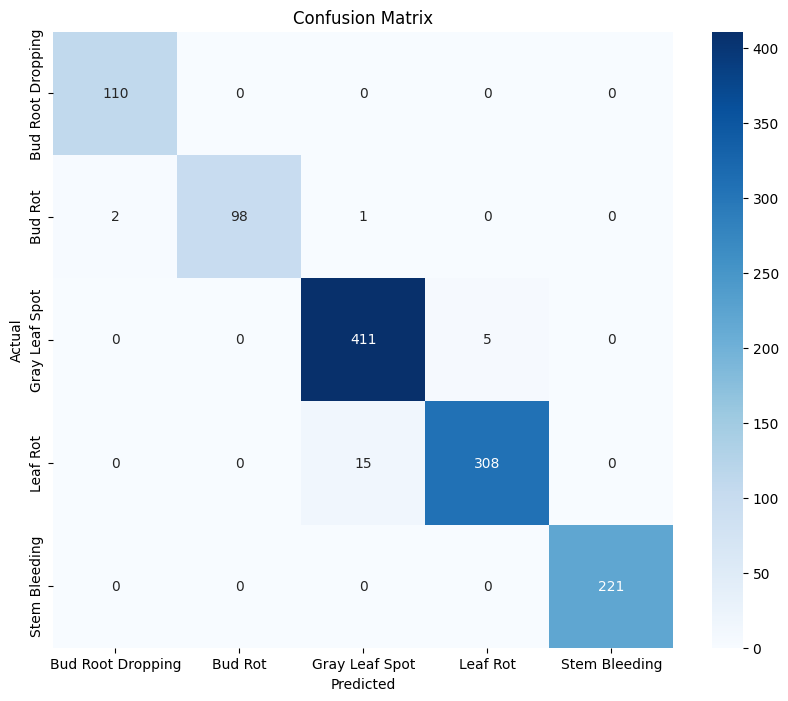}
    \caption{Confusion Matrix: EfficientNetB2 }
    \label{fig:confusion-b2}
\end{figure}

When EfficientNetB3 was used, the model accuracy increased, and therefore it was used as a feature extractor. The next subsections discuss in detail the experimental analysis using this version of EfficientNet, which is B3.
\subsubsection{Learning with Adam Optimizer}

The proposed DeepSeqCoco model was compiled with the help of Adam optimizer and was made to learn the parameters of the training data set for five epochs. The training data set was used to fit the model, and the validation data set was used to confirm it. The summary of the model is shown in Table \ref{tab:model-summary}. There were totally 10791213 parameters out of which only 7685 were considered as trainable parameters by the proposed model. 
\begin{table}[!h]
\centering
\caption{Summary of the Proposed DeepSeqCoco Model}
\label{tab:model-summary}
\begin{tabular}{lll}
\hline
\multicolumn{1}{c}{\textbf{Layer Type}}                 & \multicolumn{1}{c}{\textbf{Output Shape}}                 & \multicolumn{1}{c}{\textbf{Paramaters}}                 \\ \hline
keras\_layer                                            & (None, 1536)                                              & 10783528                                                \\ \hline
dense                                                   & (None, 5)                                                 & 7685                                                    \\ \hline
\multicolumn{3}{c}{\begin{tabular}[c]{@{}c@{}}Total params: 10791213 (41.17 MB)\\ Trainable params: 7685 (30.02 KB)\\ Non-trainable params: 10783528 (41.14 MB)\end{tabular}} \\ \hline
\end{tabular}
\end{table}

\begin{figure}[!h]
\centering
\includegraphics[width=1.00\linewidth]{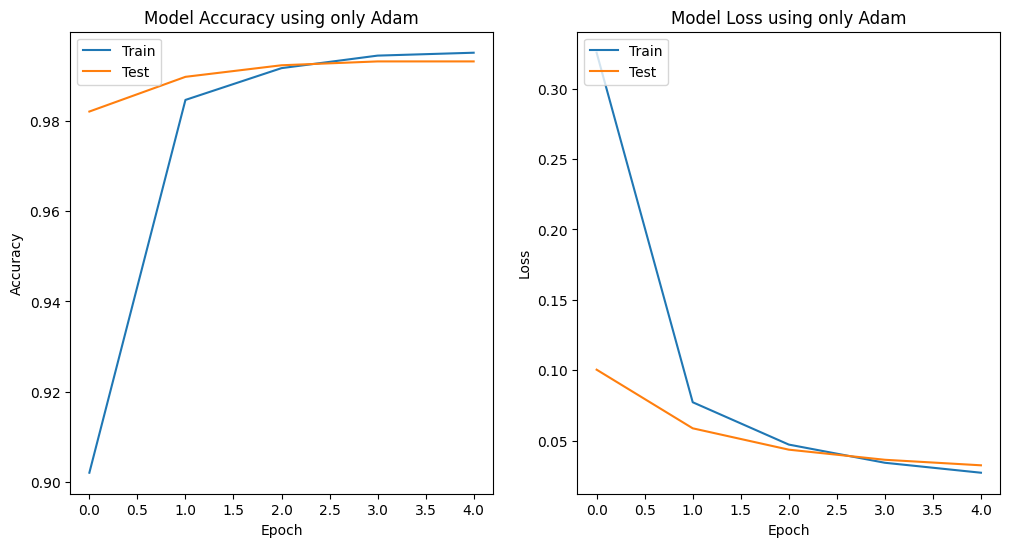}
\caption{Proposed Model Accuracy and Loss vs Number of Epochs (Adam)}
\label{fig:Adam Model Accuracy}
\end{figure}

The accuracy and loss of the model obtained during training phase was 99.5\% and 2.72\%, respectively. The time taken for 1 epoch was 1775.85 seconds. The accuracy and loss on the validation data was 99.3\% and 3.25\%, respectively. The model accuracy and loss with respect to each epoch for both training and validation data are illustrated in Figure \ref{fig:Adam Model Accuracy}. 

\begin{figure}[!h]
\centering
    \includegraphics[width=0.90\linewidth]{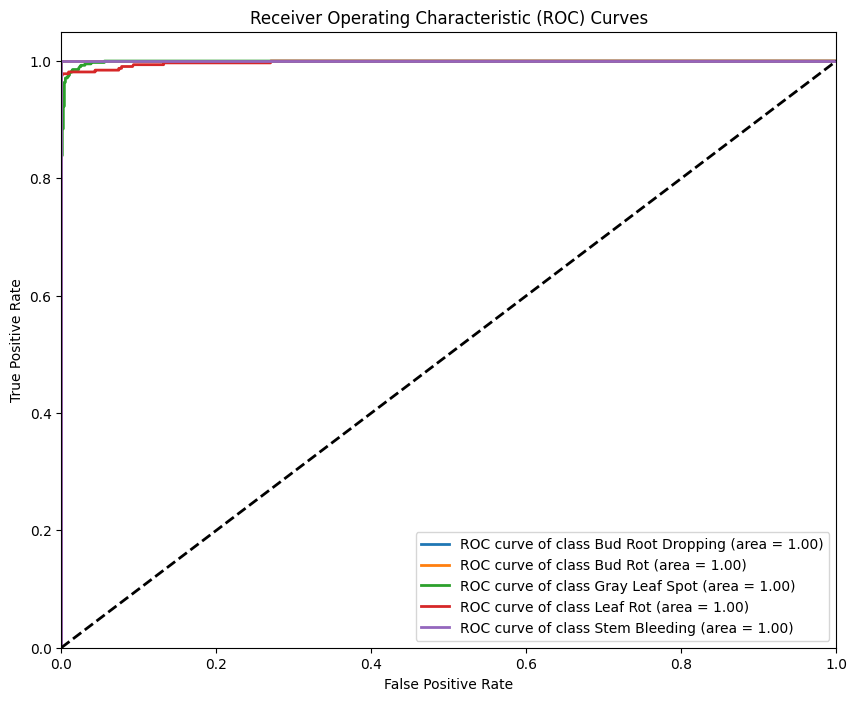}
    \caption{ROC for DeepSeqCoco Model (Adam)}
    \label{fig:roc-adam}
\end{figure}

The confusion matrix representing the total samples predicted correctly, when compared to the actual class is shown in Figure \ref{fig:confusion-adam} below. It clearly shows that only Leaf Rot and Gray Leaf Spot have false prediction. The Area Under Curve in Figure \ref{fig:roc-adam} clearly specifies that the model shows high accuracy. 
\begin{figure}[!h]
\centering
    \includegraphics[width=0.80\linewidth]{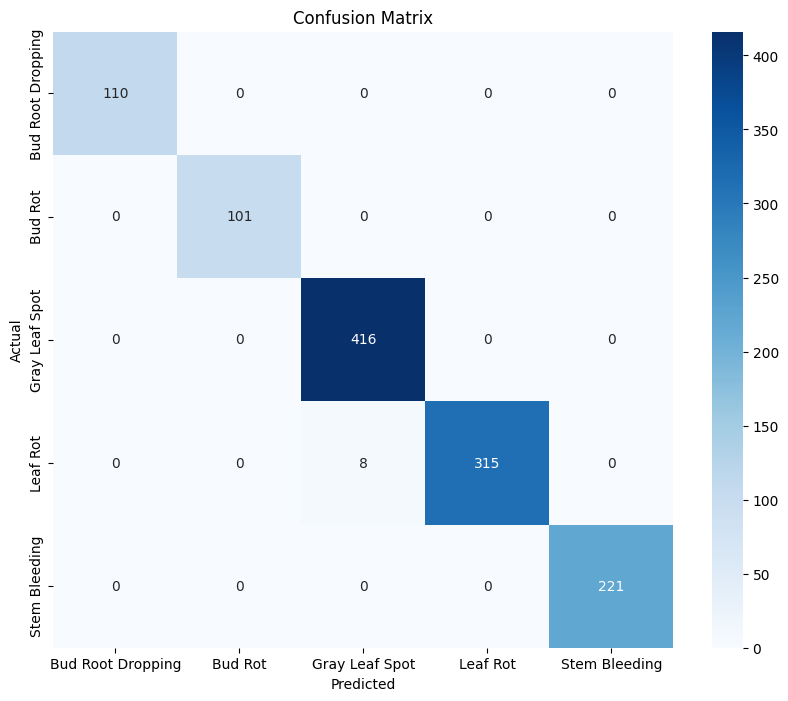}
    \caption{Confusion Matrix DeepSeqCoco Model (Adam)}
    \label{fig:confusion-adam}
\end{figure}

The other evaluation metrics which were utilized for substantiating the learning efficiency and prediction results are precision, recall, F1-score and support. The Table \ref{tab:report-class} tabulates these metrics for each class.

\begin{table}[!h]
\centering
\caption{Summary of evaluation metrics of DeepSeqCoco Model (Adam)}
\label{tab:report-class}
\begin{tabular}{|l|l|l|l|l|}
\hline
\rowcolor[HTML]{EFEFEF} 
\multicolumn{1}{|c|}{\cellcolor[HTML]{EFEFEF}\textbf{Class Name}} & \multicolumn{1}{c|}{\cellcolor[HTML]{EFEFEF}\textbf{Precision}} & \multicolumn{1}{c|}{\cellcolor[HTML]{EFEFEF}\textbf{Recall}} & \multicolumn{1}{c|}{\cellcolor[HTML]{EFEFEF}\textbf{F1-Score}} & \multicolumn{1}{c|}{\cellcolor[HTML]{EFEFEF}\textbf{Support}} \\ \hline
Bud Root Dropping                                                 & 1.00                                                            & 1.00                                                         & 1.00                                                           & 110                                                           \\ \hline
Bud Rot                                                           & 1.00                                                            & 1.00                                                         & 1.00                                                           & 101                                                           \\ \hline
Gray Leaf Spot                                                    & 0.98                                                            & 1.00                                                         & 0.99                                                           & 416                                                           \\ \hline
Leaf Rot                                                          & 1.0                                                             & 0.98                                                         & 0.99                                                           & 323                                                           \\ \hline
Stem Bleeding                                                     & 1.00                                                            & 1.00                                                         & 1.00                                                           & 221                                                           \\ \hline
\end{tabular}
\end{table}

\subsubsection{Learning with SGD Optimizer}
The total parameters for learning the training data stayed the same when the model was assembled using the SGD optimizer, setting a learning rate of 0.01 and momentum of 0.9, as indicated in Table \ref{tab:model-summary}. 
\begin{figure}[!h]
\centering
\includegraphics[width=1.00 \linewidth]{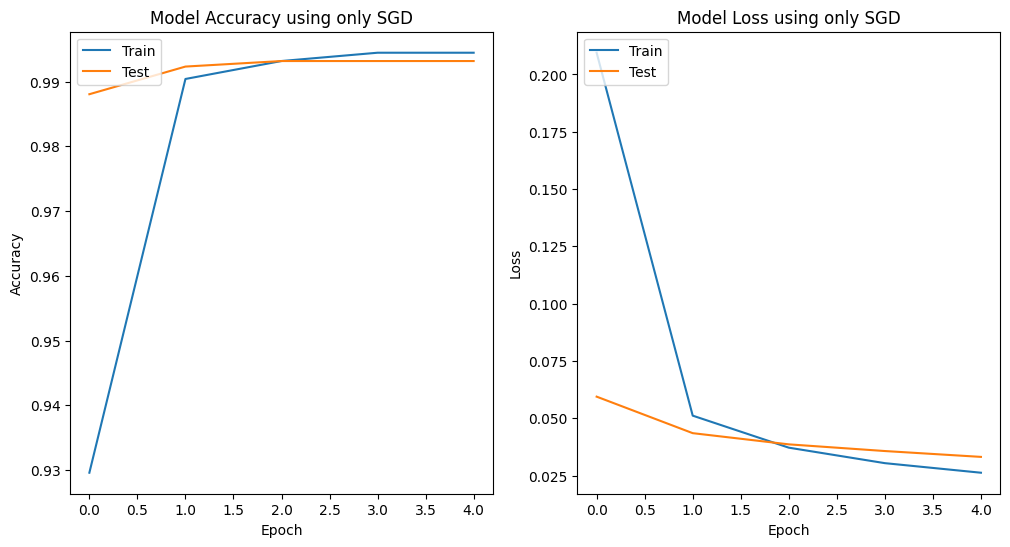}
\caption{DeepSeqCoco Model Accuracy and Loss vs Number of Epochs (SGD)}
\label{Fig:accuracy-sgd}
\end{figure}

\begin{figure}[!h]
    \centering
    \includegraphics[width=0.8\linewidth]{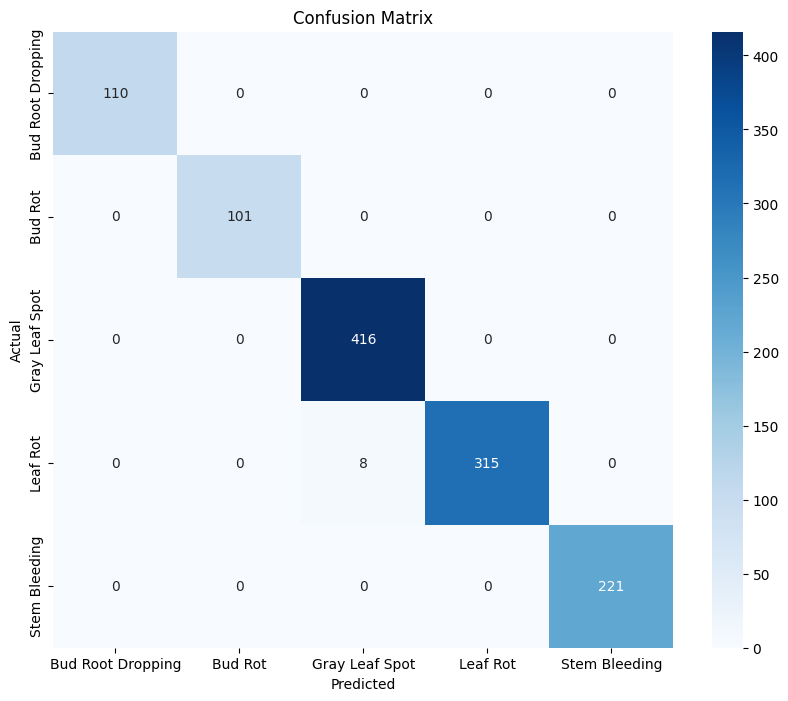}
    \caption{DeepSeqCoco Model Confusion Matrix (SGD)}
    \label{fig:confusion-sgd}
\end{figure}

The accuracy and loss obtained when the model tried to learn the parameters and predict using the training data set was 99.4\% and 2.63\%, respectively. The time taken for running 1 epoch was 1773.46 seconds. The validation accuracy and loss was 99.3\% and 3.32\%, respectively. The accuracy and loss obtained during each epoch can be known from the Figure \ref{Fig:accuracy-sgd}. The confusion matrix depicting the True Positives, False Positives, True Negatives and False Negatives are shown in the Figure \ref{fig:confusion-sgd}. Here as well, the two classes Leaf Rot and Green Leaf Spot showed variations. 

\begin{figure}[!h]
    \centering
    \includegraphics[width=0.9\linewidth]{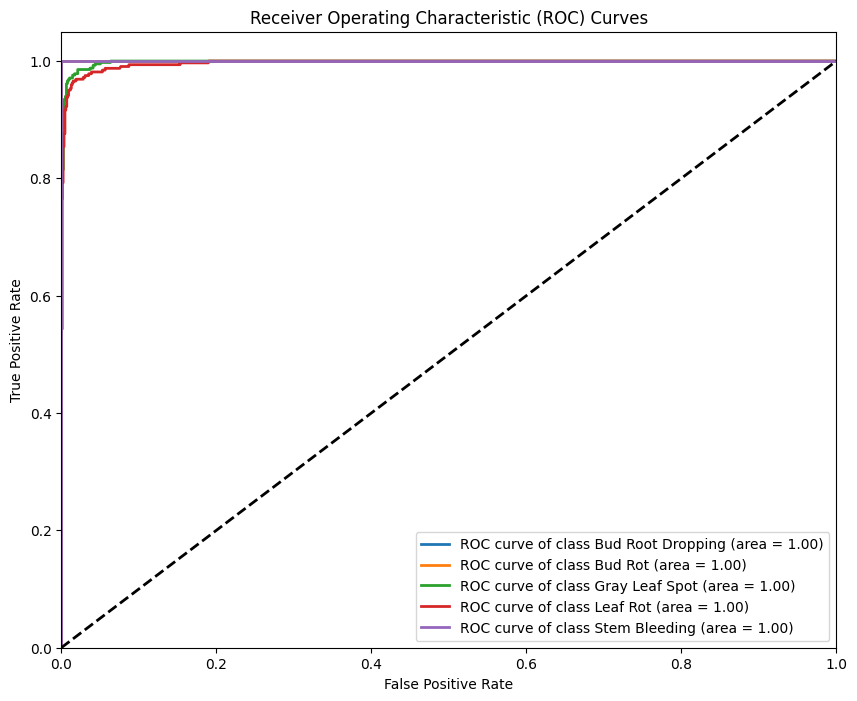}
    \caption{ROC for SGD Optimizer}
    \label{fig:roc-sgd}
\end{figure}

The RoC for the model when SGD is used is shown in the Figure \ref{fig:roc-sgd}. The various other evaluation metrics used for further analysis like precision, recall, F1-score and Support when calculated gave the same values as Adam optimizer shown in Table \ref{tab:report-class}. These parameters also specify that the two classes, Leaf Rot and Gray Leaf Spot, show lesser values 0.98 and 0.99 for F1-score, Recall and Precision.  

\subsubsection{Learning with Hybridized Optimizers}
\begin{figure}
\begin{tabular}{c}
  \includegraphics[width=80mm]{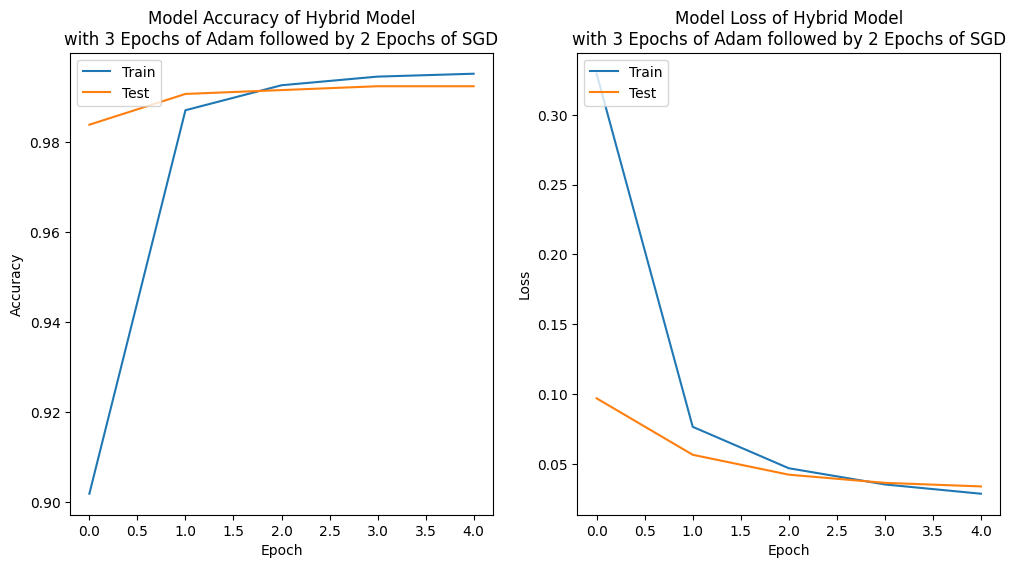}   \\
(a) Hybrid: Adam followed by SGD\\[6pt]
 \includegraphics[width=80mm]{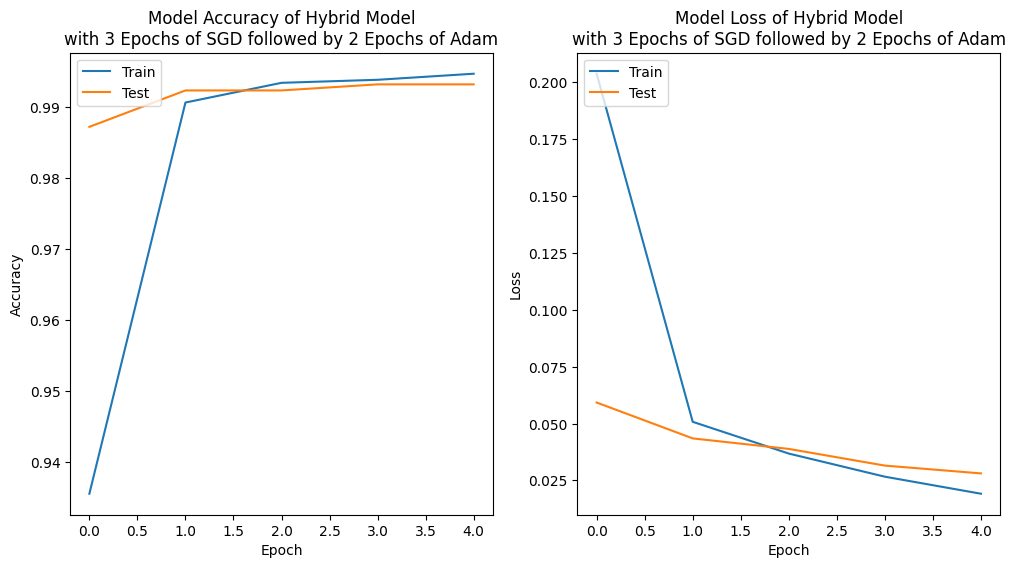} \\
(b) Hybrid: SGD followed by Adam\\[6pt]
\end{tabular}
\caption{DeepSeqCoco Model Accuracy and Loss vs Number of Epochs}
\label{Fig:Hybrid-accuracy}
\end{figure}

\begin{figure}
\centering
\begin{tabular}{c}
  \includegraphics[width=70mm]{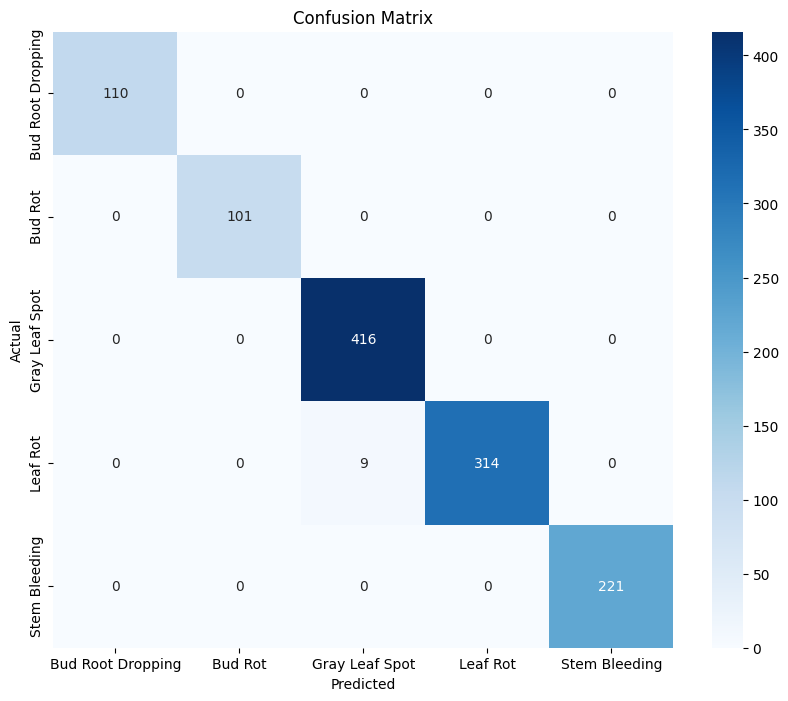}   \\
(a) Hybrid: Adam followed by SGD\\[6pt]
 \includegraphics[width=70mm]{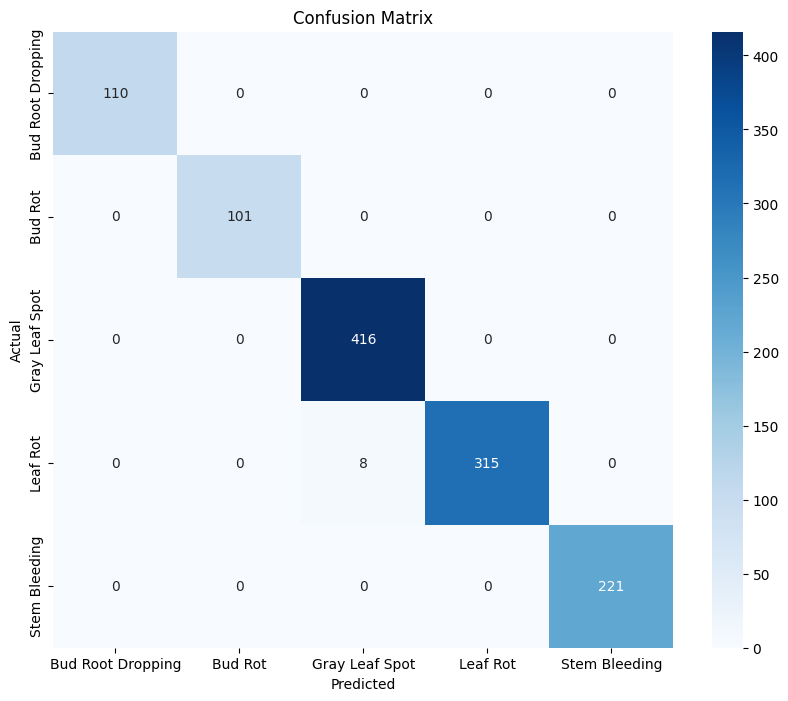} \\
(b) Hybrid: SGD followed by Adam\\[6pt]
\end{tabular}
\caption{DeepSeqCoco Model Confusion Matrix}
\label{Fig:Hybrid-confusion}
\end{figure}

The model was then compiled by hybridizing both optimizers having Adam running for 3 epochs followed by 2 epochs of SGD. Similarly, an experiment was conducted with SGD, running for 3 epochs followed by 2 epochs of Adam. The accuracy and loss obtained when these two variations were used, is shown in Figure \ref{Fig:Hybrid-accuracy}(a) and Figure \ref{Fig:Hybrid-accuracy}(b) 
respectively. When Adam was used first while hybridizing, the training accuracy and validation accuracy were 99.5\% and 99.2\%, respectively, with a training loss of 2.89\% and a validation loss of 3.41\%. Similarly, when SGD was used first, the training and validation accuracies were 99.5\% and 99.3\%, respectively, with a training loss of 1.92\% and a validation loss of 2.81\%. The overall model accuracy was 99.51\% when Adam was followed by SGD, whereas it was 99.47\%, when SGD was followed by Adam. The average training time per epoch was 1849 seconds when Adam was used first and 1804.75 seconds when SGD was used first. The confusion matrices for both cases of hybridization is shown in Figure \ref{Fig:Hybrid-confusion}. In case of Adam followed by SGD, the prediction shows an increase in false count, which is 9, when compared to the other case where it shows a count of 8.

The other evaluation metrics, when calculated, showed similar values for the hybrid SGD followed by Adam, as illustrated in Table \ref{tab:report-class}. On the other hand, the hybrid Adam followed by SGD showed mild variations, particularly in the case of Leaf Rot, where the Recall was 0.97.

\subsubsection{Comparative Analysis}
The experimental analysis clearly specifies that our proposed DeepSeqCoco model can be utilized for detecting diseases at an early stage and also with lower computational costs compared to the existing models. This section discusses the results obtained by other researchers when compared to our proposed DeepSeqCoco model, tested and validated on the same data set.  The dataset was tested using the other transfer learning models, VGG16, ResNet50, and MobileNet and the results are compared with the proposed DeepSeqCoco model.  The total parameters in case of MobileNetV2 was 2264389, VGG16 was 14717253 and ResNet50 was 23587712. When the experiment was conducted using MobileNetV2, the time required for training was 838.55 seconds per epoch and prediction time was 8.62 seconds. When VGG16 was used, the time required for learning the parameters was 1258.78 seconds per epoch and the prediction time was 4.52 seconds. The Table \ref{tab:comp} shows the comparative analysis of the computational aspects with respect to the proposed DeepSeqCoco model and other models in the literature. 
\begin{table}[!h]
\centering
\caption{Comparative Analysis of Computational Aspects}
\label{tab:comp}
\begin{tabular}{|c|c|c|c|c|}
\hline
\rowcolor[HTML]{EFEFEF} 
\textbf{Model}                                                                           & \textbf{\begin{tabular}[c]{@{}c@{}}Training \\ Time\\ (seconds)\end{tabular}} & \textbf{\begin{tabular}[c]{@{}c@{}}Prediction \\ Time\\ (seconds)\end{tabular}} & \textbf{\begin{tabular}[c]{@{}c@{}}No.of \\ Parameters\end{tabular}} & \textbf{\begin{tabular}[c]{@{}c@{}}No.of \\ Trainable\\ Parameters\end{tabular}} \\ \hline
\textbf{VGG16}                                                                           & 1258.78                                                                       & 4.52                                                                            & 14717253                                                             & 2565                                                                             \\ \hline
\textbf{ResNet50}                                                                        & 2257.85                                                                       & 13.47                                                                           & 23587712                                                             & 10245                                                                            \\ \hline
\textbf{MobileNetV2}                                                                     & 838.55                                                                        & 8.62                                                                            & 2264389                                                              & 6405                                                                             \\ \hline
\textbf{\begin{tabular}[c]{@{}c@{}}DeepSeqCoco\\ (Adam)\end{tabular}}                    & 1775.85                                                                       & 1.66                                                                            & 10791213                                                             & 7685                                                                             \\ \hline
\textbf{\begin{tabular}[c]{@{}c@{}}DeepSeqCoco\\ (SGD)\end{tabular}}                     & 1773.46                                                                       & 1.61                                                                            & 10791213                                                             & 7685                                                                             \\ \hline
\textbf{\begin{tabular}[c]{@{}c@{}}DeepSeqCoco\\ (Hybrid Adam \\ then SGD)\end{tabular}} & 1849                                                                       & 1.69                                                                            & 10791213                                                             & 7685                                                                             \\ \hline
\textbf{\begin{tabular}[c]{@{}c@{}}DeepSeqCoco\\ (Hybrid SGD \\ then Adam)\end{tabular}} & 1804.75                                                                       & 1.72                                                                            & 10791213                                                             & 7685                                                                             \\ \hline
\end{tabular}
\end{table}
From Table \ref{tab:comp}, it is clear that the proposed DeepSeqCoco model compiled with the SGD optimizer alone, predicted the disease at a faster rate for a training time of 1773.46 seconds with 7685 parameters, whereas, the longest prediction time was 13.47 seconds, when ResNet50 was used. The Table \ref{tab:validation} shows that the proposed DeepSeqCoco model performs better than the other three models in terms of accuracy, recall, precision, and F1-score.   
\begin{table}[!h]
\centering
\caption{Comparative Analysis of Evaluation Metrics}
\label{tab:validation}
\begin{tabular}{|c|c|c|c|c|}
\hline
\rowcolor[HTML]{EFEFEF} 
\textbf{Model}                                                                           & \textbf{\begin{tabular}[c]{@{}c@{}}Accuracy\\ (\%)\end{tabular}} & \textbf{\begin{tabular}[c]{@{}c@{}}Precision\\ (\%)\end{tabular}} & \textbf{\begin{tabular}[c]{@{}c@{}}Recall\\ (\%)\end{tabular}} & \textbf{\begin{tabular}[c]{@{}c@{}}F1-\\  score (\%)\end{tabular}} \\ \hline
\textbf{VGG16}                                                                           & 88                                                             & 86.94                                                           & 87.52                                                        & 87.69                                                           \\ \hline
\textbf{ResNet50}                                                                        & 94                                                             & 92.25                                                           & 93.42                                                        & 93.88                                                           \\ \hline
\textbf{MobileNetV2}                                                                     & 92                                                             & 89.68                                                           & 90.45                                                        & 91.54                                                           \\ \hline
\textbf{\begin{tabular}[c]{@{}c@{}}DeepSeqCoco\\ (Adam)\end{tabular}}                    & 99.3                                                           & 99                                                              & 99                                                           & 99                                                              \\ \hline
\textbf{\begin{tabular}[c]{@{}c@{}}DeepSeqCoco\\ (SGD)\end{tabular}}                     & 99.3                                                           & 99                                                              & 99                                                           & 99                                                              \\ \hline
\textbf{\begin{tabular}[c]{@{}c@{}}DeepSeqCoco\\ (Hybrid Adam \\ then SGD)\end{tabular}} & 99.2                                                           & 99                                                              & 99                                                           & 99                                                              \\ \hline
\textbf{\begin{tabular}[c]{@{}c@{}}DeepSeqCoco\\ (Hybrid SGD \\ then Adam)\end{tabular}} & 99.3                                                           & 99                                                              & 99                                                           & 99                                                              \\ \hline
\end{tabular}
\end{table}

\begin{figure}[!h]
    \centering
    \includegraphics[width=0.9\linewidth]{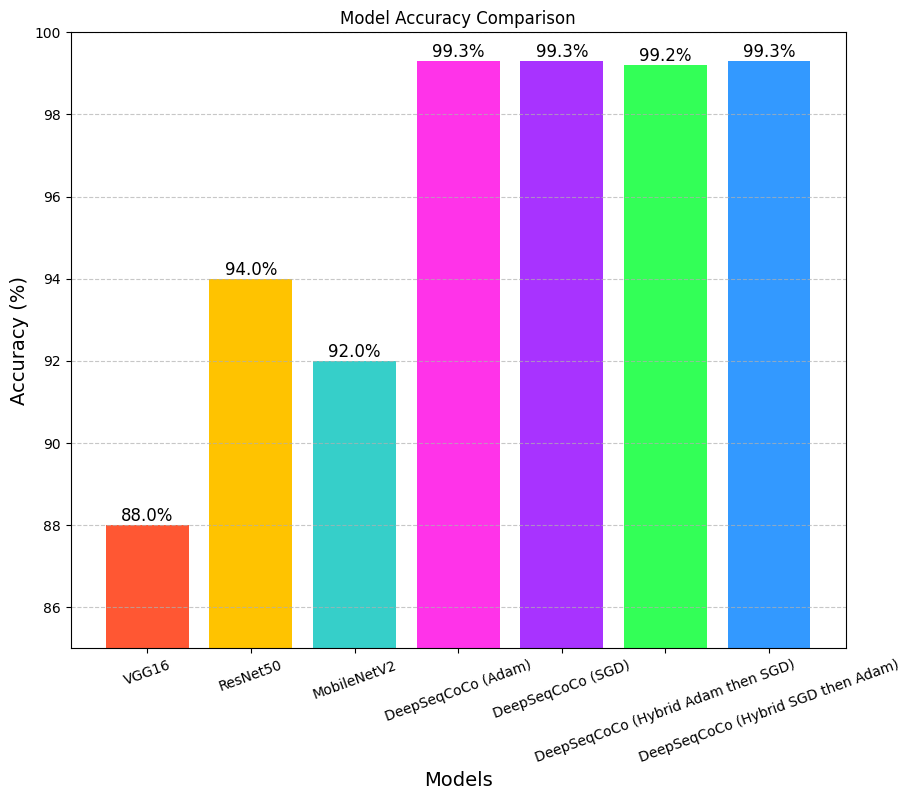}
    \caption{Accuracy of Proposed DeepSeqCoco vs other Models}
    \label{fig:validation}
\end{figure}

The validation accuracy of all variations of the proposed model is higher compared to the models discussed in the literature, as shown in Figure \ref{fig:validation}. 

The interpretations based on the experimental analysis and the results obtained as per Table \ref{tab:comp} and Table \ref{tab:validation} are as follows:
\begin{itemize}
    \item MobileNetV2 has the lowest training time (838.55 seconds), which indicates that its a lightweight architecture suitable for light-weight devices so that faster training can be done.
    \item All variations of proposed DeepSeqCoco models take longer time to train (ranging from 1773 to 1849 seconds), due to complex hybrid learning strategies as well as an increase in the number of trainable parameters.
    \item VGG16 has a relatively higher training time (1258.78 seconds) compared to MobileNetV2 but is lower than all variations of the DeepSeqCoco model.
    \item Proposed DeepSeqCoco models have the fastest prediction time ranging from 1.61 to 1.72 seconds. This parameter helps to arrive at an interpretation that this is best suited for real-time applications though the training time is slightly higher than other models.
    \item MobileNetV2 has the lowest number of parameters (2.26 million), and hence it has the lowest training time. VGG16 has a higher number of parameters (14.7 million) and hence longer training time compared to MobileNetV2. The proposed DeepSeqCoco models have totally 10.79 million parameters, but even then, they show a balance in computational costs and performance in terms of accuracy. 
    \item The proposed DeepSeqCoco models are the best choice when the requirement is for fast predictions. Also, the training time required is reasonable for increased accuracy and prediction rate. The choice of optimization strategy (Adam, SGD, or hybrid) does not have much impact on performance.
    \item The traditional model VGG16 underperformed across all metrics and hence, depicts that this is less suitable for real-time applications.  The other two models ResNet50 and MobileNetV2 showed an improved accuracy, precision and recall, and hence can be utilized in some applications where fast response is not needed. The proposed DeepSeqCoco models outperform in all metrics and hence are best suited for highly accurate and faster classification problems.

\end{itemize}
\section{Conclusion}
 In this paper, \textbf{DeepSeqCoco}, a model based on DL for efficient and reliable classification of coconut tree diseases, was proposed. Our approach improves upon existing models and methods to balance accuracy and computational efficiency which makes it more suitable for real world scenarios and use on low-power devices like mobile phones. Experimental results show that DeepSeqCoco beats traditional ML based models as well as CNN-based ones, especially when using a diverse range of images. Among the 4 variants suggested by us if computational speed is of concern, DeepSeqCoco (Adam) should be the preferred option over DeepSeqCoco (SGD), since it achieves higher training accuracy (99.5\% vs. 99.4\%) and lower validation loss (3.25\% vs. 3.32\%) with a negligible increase in training and prediction time. But if loss minimization for improved generalization is of greater concern, the hybridized model, DeepSeqCoco (SGD followed by Adam), should be employed as it has the lowest training and validation loss even though it takes a slightly longer time to train (when compared to non-hybridized models) and shows a negligible increase in prediction time.

 The timely detection of diseases is important to prevent large-scale damage to crops and to ensure sustainable farming. DeepSeqCoco contributes to this by offering a reliable and convenient solution for farmers. To improve trust, future research should concentrate on growing the dataset to include even more environmental variances. Development of a real time mobile application for disease detection should also be considered. By utilizing AI-driven solutions like DeepSeqCoco, we aim to support improved agriculture and coconut crop health at a larger scale.
\bibliography{access.bib}
\bibliographystyle{ieeetr}

\end{document}